\documentclass[11pt]{article}

\usepackage[margin=1in]{geometry}

\usepackage[T1]{fontenc}

\usepackage{lmodern,microtype}

\usepackage{amsmath,amssymb,amsthm,mathtools}

\usepackage{booktabs,tabularx,longtable,array}

\usepackage{graphicx,xcolor,tikz}

\usetikzlibrary{arrows.meta,positioning,calc,fit}

\usepackage{enumitem,fancyvrb}

\usepackage{xurl}

\usepackage[colorlinks=true,linkcolor=blue!45!black,citecolor=blue!45!black,urlcolor=blue!45!black]{hyperref}

\usepackage[nameinlink,noabbrev]{cleveref}

\usepackage{caption}

\setlist{itemsep=0.2em,topsep=0.4em}

\newcolumntype{Y}{>{\raggedright\arraybackslash}X}

\newtheorem{theorem}{Theorem}[section]

\newtheorem{lemma}[theorem]{Lemma}

\newtheorem{corollary}[theorem]{Corollary}

\theoremstyle{definition}

\theoremstyle{remark}

\DeclareMathOperator{\rank}{rank}

\DeclareMathOperator{\diag}{diag}

\DeclareMathOperator{\adj}{adj}

\newcommand{\Q}{\mathbb{Q}}

\newcommand{\C}{\mathbb{C}}

\newcommand{\VP}{\mathrm{VP}}

\newcommand{\VNP}{\mathrm{VNP}}

\newcommand{\code}[1]{\nolinkurl{#1}}

\hypersetup{pdftitle={Recursive Reasoning Exploration: Mathematical Research through Evolving Reasoning Graphs. case study: An Exact Certificate-Degree Theorem for Two-Term Tensor Decompositions},pdfauthor={Ying Xie},pdfsubject={Phase III of the Math Instruments Program; exact certificate degree for two-term tensor decompositions}}

\title{Recursive Reasoning Exploration:\\Mathematical Research through\\Evolving Reasoning Graphs\\[0.6em]\large case study: An Exact Certificate-Degree Theorem\\for Two-Term Tensor Decompositions\\[0.45em]\normalsize Phase III of the Math Instruments Program}

\author{Ying Xie\\Kennesaw State University\\\texttt{yxie2@kennesaw.edu}}

\date{}

\begin{document}

\maketitle

\begin{abstract}

Large collections of mathematical manuscripts and research notes create an opportunity to reuse arguments at a scale that exceeds individual reading, while increasing the difficulty of retaining hypotheses, checking claims, and understanding how results were obtained. We propose Recursive Reasoning Exploration (RRE), a framework for cumulative mathematical research through evolving reasoning graphs. RRE links claims and jointly required premises to research targets, attempted arguments, counterexamples, checks, and revisions. Each investigation can generate further targets, and its supported outcomes become available to subsequent investigations. This is Phase III of the Math Instruments program: identifying mathematical instruments, reconstructing reusable reasoning, and organizing recursive research from that reasoning. The framework supports human-directed, collaborative, and goal-directed autonomous exploration, with open-ended exploration as a further mode. Its records also provide material for learning how mathematical investigations develop. A complexity research case study establishes an exact theorem: for every complex tensor of order $n\ge2$ and rank greater than two, the minimum Nullstellensatz certificate degree for its original two-term decomposition equations is $3n$. This includes an infinite family whose every flattening has rank at most two. The exploration develops an explicit boundary certificate, a balanced transfer lemma, and an exhaustive argument using classical rank-two geometry. A parallel branch gives bounds $3m+3\le\delta_m(I_m,N)\le3m+6$ for every nonzero nilpotent $m\times m$ matrix. Six graph revisions record the progression from inherited results to these theorems, with exact arithmetic checks and a replayable exploration history. We present the RRE representation and protocol, the mathematical proofs, visualizations of graph evolution, and a research agenda for shared human--AI exploration, cumulative reuse, and education.

\end{abstract}


\clearpage

\section{From mathematical abundance to cumulative research}\label{sec:intro}

Mathematical research produces more than the statements eventually published. It produces representations, intermediate lemmas, tests of plausible routes, failures that reveal necessary hypotheses, and decisions about which question to pursue next. Much of this work remains in notebooks, conversations, drafts, or personal memory. A completed paper selects a coherent argument from that process. This selection is essential for communication, but it can leave later researchers unable to tell why an approach was attempted or which nearby routes have already been tested. When the exploration record is unavailable, both human and AI investigators may spend effort repeating unsuccessful approaches. Such a record is useful only when it preserves the target, assumptions, method, and outcome: an unsuccessful attempt is not by itself a proof that a route is impossible.

AI-assisted mathematical work makes this issue more pressing. Published papers and manuscripts may present a result and its polished proof without preserving the external research history of targets, candidate arguments, tests, and revisions that led to them. An inspectable record of that history could help experts assess how a result was reached and give students material for understanding research decisions, not merely its final proof. The record documents observable research actions and evidence. A large release of manuscripts can contain useful mechanisms even when individual claims require correction or further scrutiny. It can also create a substantial review burden. The quantity of generated text gives little information about mathematical validity, novelty, explanatory value, or usefulness to another researcher. These are different questions, and infrastructure should preserve their differences. The Leiden Declaration on Artificial Intelligence and Mathematics emphasizes human understanding, responsibility, and the needs of reviewers \cite{leiden}. A research system that produces more material without making its assumptions and evidential basis easier to inspect may worsen the problem it aims to solve.

The central proposal of this paper is to make the \emph{research investigation} a reusable mathematical object. Its record should connect a target to the arguments that motivated it, the conditions required for progress, the outcomes of attempted routes, and the changes those outcomes make to subsequent research. By making supported results and carefully scoped unsuccessful attempts available to later investigations, RRE aims to reduce redundant effort and help research accumulate across people and systems. The proposed unit of accumulation is therefore larger than a theorem and more structured than an unedited conversation. It includes a claim, its justification, its scope, its provenance, and the research decisions that made it relevant.

\subsection{Three phases of the Math Instruments program}

Phase I identifies reusable mathematical instruments in research manuscripts \cite{xie1}. An instrument may be a construction, a proof mechanism, or a way to introduce useful structure into a problem. The motivating survey examined an initial collection of 722 AI-written manuscripts and selected 38 problem families for closer study. Its purpose was to identify potentially reusable ideas while retaining uncertainty about validity, antecedents, and novelty.

Phase II reconstructs the reasoning that makes such instruments usable \cite{xie2}. It represents claims and their jointly required premises through source-linked directed hypergraphs, preserving hypotheses, evidence, and transfer obligations. Its six-paper study consists of four OpenAI manuscripts and two retrospective application papers by the author. The study reports 158 arguments and 454 inferences, together with a multiscale visualization. Phase II also proposes a research cycle in which a curated collection supports candidate approaches, hypothesis suggestions, transfer investigations, verification, and the return of outcomes to the collection.

Phase III develops that proposed cycle into \emph{Recursive Reasoning Exploration}. The emphasis moves from extracting an argument to conducting and recording an investigation that can repeatedly extend the argument collection. Building on Phase II, it specifies how these ideas operate together: explicit objectives, generated intermediate targets, recursive investigation, evidence-sensitive graph updates, and declared stopping criteria.

\begin{table}[htbp]
\centering\small
\begin{tabularx}{\linewidth}{@{}p{1.35cm}YYY@{}}
\toprule
Phase & Primary question & Main record & Next use \\
\midrule
I & What mathematical device may be reusable? & Instrument, source, antecedents, caveats & Select mechanisms worth reconstructing \\
II & Why does it work, and under which hypotheses? & Reasoning graph with scope and transfer obligations & Find candidate routes and missing premises \\
III & What should be investigated next, and how does the outcome change what we know? & Evolving reasoning graph and exploration history & Reuse results and failures in subsequent research \\
\bottomrule
\end{tabularx}
\caption{The phases describe a research program. They are not levels of verification: every phase needs its own mathematical review.}
\label{tab:phases}
\end{table}

\subsection{What Recursive Reasoning Exploration means}

RRE combines a collection of reasoning graphs with research objectives, a resource budget, and criteria for accepting results. Objectives may be supplied by researchers or formulated during exploration under a declared research policy. For each investigation, RRE identifies an unresolved obligation, retrieves relevant arguments, proposes a route, and investigates the additional claims that route requires. A result can resolve the target, qualify it, refute a proposed statement, or leave it open. The record is updated accordingly, and the updated record helps determine the next investigation.

The word \emph{recursive} has two meanings. Within an investigation, a proposed argument generates subproblems that are investigated by the same protocol. Across investigations, an outcome changes the available knowledge and can generate a new target. The word \emph{reasoning} identifies the shared object: explicit mathematical arguments that people and AI systems can inspect, revise, and extend.

We distinguish four operating modes according to how objectives are formulated and investigations are conducted (\cref{tab:operating-modes}). The second mode, \emph{Collaborative Reasoning Exploration} (CRE), is also called collaborative RRE. The fourth mode is prospective: an agent would examine the collection, formulate research objectives, propose connections or mathematical constructions, and investigate their consequences without receiving a specific mathematical problem for each investigation. It would still operate under explicit criteria for mathematical value, evidence, and resource use.

\begin{table}[htbp]
\centering\small
\begin{tabularx}{\linewidth}{@{}>{\raggedright\arraybackslash}p{4.1cm}Y@{}}
\toprule
Operating mode & Objective formation and investigation \\
\midrule
Human-directed exploration & Researchers choose and investigate questions, using the graph to organize their work. \\
Collaborative Reasoning Exploration & Humans and AI jointly formulate questions, investigate arguments, and extend the graph, with recorded contributions and handoffs. \\
Goal-directed autonomous exploration & A researcher supplies the overall objective; an AI agent recursively selects and investigates intermediate targets within the declared charter. \\
Open-ended autonomous exploration & An AI agent formulates research objectives from the collection and investigates them under standing evaluation and resource policies. \\
\bottomrule
\end{tabularx}
\caption{Four operating modes of RRE. They describe the allocation of research activities, not levels of mathematical verification. The case study illustrates the third mode.}
\label{tab:operating-modes}
\end{table}

A project can move between modes while preserving one graph and history. Automation does not permit self-certification or relax the rules for admitting a mathematical claim. \emph{Research-strategy improvement} is a separate proposed capability: the system could evaluate and revise its own retrieval, target-selection, or investigation methods across these modes. Adding graph records alone does not establish such an improvement. The case study uses the run identifier RRE005 and illustrates goal-directed autonomous RRE.

Explainability here has an operational meaning. A reader should be able to inspect which premises support a conclusion, why a target was selected, which alternatives were recorded, what was checked, and what remains unresolved. These records are mathematical justifications and externally inspectable research actions. They are not a claim to recover a model's internal computations or a mathematician's unrecorded thought process.

The intended outcome is a shared environment in which human researchers and AI systems can contribute to the same evolving mathematical record. An expert may formulate a question, an AI system may propose a lemma, a computer algebra program may test an identity, and another researcher may discover a counterexample. The record should preserve those different contributions without collapsing them into a single label such as ``AI-generated proof.''

\subsection{Contributions and case-study progression}

This paper makes three connected contributions. First, it specifies RRE as a protocol connecting reasoning graphs, research targets, verification records, and versioned exploration events. Second, it develops a substantive mathematical case study: an exact certificate-degree theorem for all inconsistent two-term tensor decomposition systems, together with nearly sharp bounds for nilpotent matrix pencils. Third, it presents a research agenda for shared human--AI exploration, educational access to research paths, and evaluation of cumulative progress.

The case study begins with two results already in the complexity atlas: a universal lower bound on certificate degree and a matching upper bound when a flattening minor detects excessive rank. The author selected the missing boundary case as a new objective: obtain matching bounds on an infinite family that flattenings miss. The agent first developed exact small certificates, then a uniform argument for nilpotent pencils. The first certificate supplied a concrete base for another target: transfer it to higher tensor orders while charging the contracted modes in the original parameters. Balance was already available from symmetry reduction; the useful additional step was to show that a balanced degree block of size $k$ transfers at cost $n$. The resulting lemma, combined with classical rank-two geometry, gives the exact value $3n$ for every order-$n$ tensor of rank greater than two. These dependencies and the separate historical sequence are the central example of RRE in this paper.

\Cref{sec:case} states the theorem and follows the exploration through six saved graph revisions. \Cref{app:math} supplies the full proofs and the explicit base certificate. \Cref{app:records} documents the event history, exact checks, and replay package; \cref{app:complexity-graphs} explains how the underlying complexity atlas was extracted and how the new component was integrated. The author supplied the objective and milestone; the AI research agent developed the recorded arguments, programs, and subsequent targets. The study documents a mathematical outcome and its research process. Comparative evaluation of the framework is developed in \cref{sec:evaluation}.

\subsection{Intellectual context and related systems}

The idea of making mathematical practice explicit has substantial antecedents. P\'olya's treatment of problem solving emphasizes questions and heuristics that guide a search \cite{polya}. Lakatos examines how conjectures and proofs develop through criticism and revision \cite{lakatos}. RRE's retention of failed routes and scope corrections belongs to this tradition.

Proof planning represents strategies and their applicability conditions to guide theorem proving \cite{bundy}. Modern neural proof search, including HyperTree Proof Search, already uses structured search and learning from earlier proof attempts \cite{htps}. RRE's AND-premise representation and recursive subgoals therefore have clear precedents. Its proposed emphasis is a persistent research record spanning problem formulation, informal and formal arguments, quantitative feasibility, failed transfers, review, and reuse across investigations. A proof-search engine could be one component within that record.

Mathematical knowledge management provides another foundation. MMT and related work address modular theory graphs, large interconnected corpora, and versioned mathematical knowledge \cite{mkm}. The Lean mathematical library demonstrates community development of a reusable formal library \cite{mathlib}. RRE should connect to such systems rather than introduce an incompatible replacement for formal semantics or proof checking. An informal reasoning graph can identify material worth formalizing; it does not inherit kernel verification merely from pointing toward a proof assistant.

Polymath provides a precedent for distributed mathematical collaboration with visible contributions \cite{polymath}. Provenance standards such as W3C PROV describe entities, activities, and agents involved in producing research artifacts \cite{prov}. RRE can reuse these ideas while adding mathematical scope, obligations, and evidence types. A chronology of contributions and a dependency graph answer different questions and should remain linked but distinguishable.

AI-driven scientific exploration also predates this proposal. FunSearch combines program generation, evaluation, and a population of retained programs \cite{funsearch}; The AI Scientist describes an iterative workflow extending from ideas to experiments and reports \cite{aiscientist}. RRE focuses on mathematical argument reuse and the explicit treatment of open obligations, counterexamples, and verification boundaries. The contribution here is their integration around mathematical dependencies, quantitative obligations, and an evolving research record.

The mathematical case study also connects to the geometry and complexity of tensor decomposition. Flattening equations for the second secant variety were established by Landsberg and Manivel and strengthened by Raicu \cite{landsbergmanivel,raicu}; Ballico and Bernardi describe ranks and decompositions on the tangent developable \cite{ballicobernardi}. Aliabadi and Friedland formulate tensor-rank testing through polynomial equations and effective Nullstellensatz linear systems \cite{aliabadifriedland}. Our result determines the exact certificate degree in the original two-term parameters, including tensors of border rank two whose decomposition fibers are empty. Algebraic proof complexity for tensor isomorphism studies a related but different equation system \cite{galesi}. The distinction between those objects guides both the theorem and its graph representation.

\section{A research state with mathematical and historical structure}\label{sec:representation}

\subsection{Reasoning graphs and exploration histories}

An RRE state at revision $t$ is represented schematically as
\[
  \mathcal S_t=(\mathcal G_t,\mathcal H_t,\mathcal F_t,\mathcal A_t,\mathcal P_t),
\]
where $\mathcal G_t$ is the reasoning collection, $\mathcal H_t$ the exploration history, $\mathcal F_t$ the frontier of research targets, $\mathcal A_t$ the evidence and artifact store, and $\mathcal P_t$ the active objectives, operating mode, budget, and research and admission policies. Each objective records its origin, selection rationale, and acceptance conditions. In open-ended exploration, new objectives become explicit records before their investigation; subsequent revisions preserve their earlier versions.

The reasoning collection answers: \emph{What supports this mathematical statement?} The exploration history answers: \emph{What was attempted, when, using which available evidence, and with what outcome?} A proof may be reorganized after discovery into a shorter dependency graph. That reorganization should not rewrite the historical sequence of investigations. Conversely, an event that motivated a proof step is not necessarily a mathematical premise of that step.

The graph has typed nodes for definitions, assumptions, claims, constructions, counterexamples, obligations, and evidence records. An inference is a directed hyperedge
\begin{equation}\label{eq:hyperedge}
 e:\quad \Gamma_e;\ p_1,\ldots,p_k\ \Longrightarrow\ q,
\end{equation}
with all $p_i$ jointly required, an explicit scope $\Gamma_e$, and a linked justification. Alternative arguments for $q$ are separate hyperedges. The existence of an edge records an asserted inference, not automatic mathematical truth. Its status depends on the justification and on the statuses of its premises.

Scope includes domains, quantifiers, parameter dependence, and limiting regimes. A theorem for each fixed depth is different from a uniform theorem over growing depth. A result valid on a generic set is different from one valid at every parameter value. These distinctions are part of the mathematical content, not optional descriptive metadata. Where a common condition is inherited through an argument, a prerequisite trace must still expose it.

Navigation links record similarity, analogy, source relationships, or historical motivation. They never establish implication. This distinction is particularly important in an aggregated graph: two papers may share an instrument without one theorem supplying the hypotheses of the other. A visually short path is not necessarily a valid mathematical route.

\subsection{A glossary of status and review}

RRE separates four questions: what kind of mathematical assertion is recorded, what has been checked, how significant it might be, and whether it is new. A single ordered confidence scale cannot represent these distinctions adequately.

\begin{table}[htbp]
\centering\small
\begin{tabularx}{\linewidth}{@{}p{2.7cm}Y@{}}
\toprule
Mathematical status & Meaning in this paper \\
\midrule
Source assertion & A statement faithfully attributed to a source; independent validity is not implied. \\
Conjecture & A proposed statement without an admitted proof. \\
Conditional derivation & An argument with named assumptions or unresolved premises still attached. \\
Proof recorded & A complete argument is supplied at its declared level of rigor; its review record remains separate. \\
Computation & An exact or numerical result for stated inputs or a finite range. It supports only the claim actually checked. \\
Refutation & A counterexample or argument contradicts a specified statement under the same hypotheses. \\
Open obligation & A claim or quantitative bound is required by a route and has not been established. \\
\bottomrule
\end{tabularx}
\caption{Mathematical statuses. Supersession and withdrawal are revision states; they preserve the earlier record. A failed attempt is an event outcome, not a refutation.}
\label{tab:statuses}
\end{table}

Review is a collection of scoped records: source comparison, internal AI review, exact certificate replay, independent implementation, human review, or formal verification. These are not interchangeable. Exact arithmetic can certify a finite identity without establishing an all-size theorem. A formal kernel can validate an encoded theorem while leaving the correspondence to an informal source in need of review. Human review should identify who reviewed what, with which limitations; responsibility for a manuscript does not establish that every intermediate claim has been independently checked.

Imported mathematical results retain their attribution and assumptions. A classical theorem used as an explicit import is an ordinary mathematical dependency. For compatibility with Phase II, a \emph{source dependency} means an unresolved reliance on a particular extracted source assertion or certificate contract, rather than every imported theorem. This convention prevents a label intended to flag an unreviewed research input from also flagging routine uses of established linear algebra.

Novelty fields distinguish ``known in the collection,'' ``new to the selected record,'' and ``literature novelty unresolved.'' A search that fails to locate a prior statement does not prove novelty. Significance is assessed separately in \cref{sec:significance}. The case study attributes inherited lemmas and classical inputs separately from the certificate construction and transfer developed during the run.

\subsection{Conditional support and safe graph updates}

Consider a proposed route
\[
 A\Longrightarrow B,\qquad B\land C\Longrightarrow D.
\]
The route reaches $D$ only after $C$ is established and the two occurrences of $B$ have compatible meanings and scopes. If $C$ is merely conjectured, the graph may record a conditional argument for $D$ but cannot silently mark $D$ as proved. A counterexample to one proposed $C$ eliminates that route in its stated scope, not every possible proof of $D$.

An update therefore has two effects. It appends historical records and changes the current view of supported routes. Historical storage can be append-only while current admissibility changes nonmonotonically. If a premise is withdrawn, affected descendants are marked for reassessment. A conclusion with an alternative valid derivation may remain supported. The system must assess individual derivations rather than indiscriminately invalidate every connected node.

For ordinary finite arguments, the admitted proof dependencies should be acyclic. Exploration can revisit a target many times, and the historical graph can contain feedback patterns. Such feedback cannot establish two unsupported claims by having them cite each other. Mathematical induction or a justified fixed-point argument is represented through its explicit theorem and discharge conditions, not through an unexplained cycle of assertions.

Structural validators can detect unresolved identifiers, missing evidence files, malformed scopes, or circular dependencies. They cannot generally decide whether an informal inference is valid or whether two hypotheses are equivalent. The graph is a means of exposing these questions to tools and reviewers; its structural validity is a necessary bookkeeping property, not a proof certificate for the mathematics.

\subsection{The research target as a first-class record}

Each target specifies a statement, its parameter regime, the objective-relevant gap it addresses, candidate source arguments, an acceptance criterion, and the consequences that success would enable. It also records an investigation budget and the evidence available when the target was selected. An anticipated downstream implication must itself be justified or listed as conditional.

For example, the target in the case study asks for matching upper and lower bounds on the degree of a polynomial identity in the original tensor-decomposition parameters. A proof using inverses or normalized variables must account for the cost of returning to that parameter model. The target specifies an infinite family, so a certificate for one small tensor supplies a new input rather than completing the objective. Recording these requirements makes the next useful question explicit before significance is assigned.

The target frontier need not contain only theorem statements. It can include a request to test a proposed mechanism against a known obstruction, compare two quantifier orders, formalize a critical lemma, or reproduce a computational certificate. These activities have research value when their consequences for the objective are explicit.

\section{The recursive exploration protocol}\label{sec:protocol}

\subsection{Initialization and target generation}

A run begins by recording its operating mode, initial graph revision, admissible sources, resource budget, and stopping criteria. Goal-directed exploration also pins the overall objective and operational milestone. Open-ended exploration instead specifies a policy for generating and selecting mathematical objectives, such as resolving recorded obstructions, connecting compatible arguments, or developing reusable constructions. Each selected objective and its acceptance conditions are recorded before investigation. An ambitious objective can coexist with a modest operational milestone. A milestone must specify a mathematical or infrastructural deliverable, and changes to its acceptance conditions become explicit events.

Retrieval begins with an active objective or an examination of the collection's open obligations under the objective-generation policy. Candidate connections can arise by composing arguments, transferring a construction to a new setting, strengthening a quantitative bound, weakening a hypothesis, or investigating why a recorded counterexample defeats a route. A system can also propose a conjecture before a researcher has specified a subproblem. The proposal should include a candidate reasoning graph with its missing premises, rather than just a plausible sentence.

Semantic similarity serves as a retrieval aid. Mathematical compatibility requires further work: identify the objects, map the hypotheses, track constants and quantifiers, and show that the proposed conclusion would be useful. A transfer proposal is admitted as a candidate until these bridge obligations are discharged. This makes automatically suggested hypotheses inspectable without presenting them as established results.

\begin{table}[htbp]
\centering\small
\begin{tabularx}{\linewidth}{@{}p{2.6cm}YY@{}}
\toprule
Operation & Question generated & Admission obligation \\
\midrule
Composition & Can one result supply a premise of another? & Align objects, scopes, and quantifiers \\
Transfer & Can a mechanism work in a different setting? & Prove the correspondence and missing bridge lemmas \\
Quantitative refinement & Can a bound meet the next theorem's requirements? & Retain every growing parameter and cost \\
Generalization & Which assumption can be weakened? & Check the modified proof and boundary cases \\
Obstruction analysis & What condition avoids a counterexample? & State exactly what the counterexample rules out \\
Structural synthesis & Do several arguments express one reusable mechanism? & Separate analogy from a proved common statement \\
\bottomrule
\end{tabularx}
\caption{Target-generation operations. They organize candidate questions; none guarantees a useful discovery.}
\end{table}

\subsection{Selection, investigation, and feedback}

Targets are selected according to their relevance to the objective, the obligations they could resolve, feasibility, estimated resource cost, and duplication of earlier work. These are scheduling judgments. A high priority is not a probability that a statement is true. A practical scheduler should retain several different approaches and revisit neglected targets when new evidence changes their prospects.

Investigation alternates constructive and adversarial work. Constructive work develops a proof, representation, algorithm, or example. Adversarial work tests endpoint cases, hidden uniformity assumptions, degenerate parameters, counterexamples in the collection, and possible loss of hypotheses. If a proof requires a new lemma, that lemma becomes a child target. Child targets inherit the parent deadline and the conditions needed by the parent argument.

An investigation can terminate with a proof, a conditional result, a counterexample, a finite computation, a failed attempt, or an unresolved question. Each outcome has a different graph effect. A proof supplies a new candidate inference with evidence. A counterexample blocks a specific statement or route. A failed attempt records a method that did not succeed; it does not prove impossibility. An expired budget produces an open obligation, not a negative theorem.

After review, the graph is updated and affected targets are reconsidered. A new lemma may satisfy a premise of an older route. A counterexample may suggest a weaker statement. A quantitative limitation may redirect attention from one resource to another. In the case study, a finite boundary certificate first motivates a uniform pencil bound. A subsequent target asks how to charge contracted modes when transferring certificates to higher tensor orders. Symmetry reduction already supplies balanced multipliers; the transfer argument establishes the degree factor $n/k$. Classical geometry then extends its application to the complete two-term model.

\begin{figure}[htbp]
\centering
\begin{tikzpicture}[>=Latex,font=\small,
 box/.style={draw=blue!50!black,rounded corners=2pt,fill=blue!4,align=center,text width=4.4cm,minimum height=1.55cm,inner sep=7pt},
 arr/.style={->,thick,draw=blue!50!black}]
\node[box] (objective) at (0,0) {Research policy and current graph\\\footnotesize objectives, scopes, open obligations};
\node[box] (target) at (5.7,0) {Formulate an objective\\or target\\\footnotesize candidate route and admission bounds};
\node[box] (investigate) at (5.7,-2.7) {Investigate and test\\\footnotesize prove, compute, challenge; create subtargets};
\node[box] (review) at (0,-2.7) {Review the outcome\\\footnotesize evidence, assumptions, limits};
\node[box] (update) at (0,-5.4) {Append the graph change\\\footnotesize supported result, refutation, or open obligation};
\node[box] (next) at (5.7,-5.4) {Reassess the frontier\\\footnotesize new questions; milestone or budget check};
\draw[arr] (objective)--(target);
\draw[arr] (target)--(investigate);
\draw[arr] (investigate)--(review);
\draw[arr] (review)--(update);
\draw[arr] (update)--(next);
\draw[arr] (next.east)--++(0.5,0)|-([yshift=0.65cm]target.north)--(target.north);
\node[align=center,font=\footnotesize,text width=10.7cm] at (2.85,-7.4)
 {Human researchers and AI systems can contribute at each stage.\\The event history links the state, action, evidence, and resulting revision.};
\end{tikzpicture}
\caption{The proposed RRE cycle. Arrows describe research activity, not mathematical implication. Feedback may generate new targets and, in the open-ended mode, new overall objectives under the declared policy; it does not permit circular proofs. A stopping decision saves the state and remaining obligations.}
\label{fig:cycle}
\end{figure}

\subsection{A controller specification}

The following pseudocode specifies behavior rather than a particular software architecture. ``Review'' includes source and mathematical assessment at a declared level; it is not assumed to be an infallible algorithm.

\begin{quote}\small
\begin{enumerate}[leftmargin=1.8em]
\item Pin the operating mode, graph revision, source versions, review policy, and deadline, together with an objective and milestone or an objective-generation policy.
\item Record the active objectives and build a frontier from open obligations and candidate connections.
\item While time remains and no stopping condition is met:
  \begin{enumerate}[label=\alph*.,leftmargin=1.6em]
  \item Select a target; record its inputs, rationale, and allocation.
  \item Investigate it; create child targets for missing steps and test the proposed route against relevant obstructions.
  \item Save the outcome, argument, computations, and unresolved obligations.
  \item Review the outcome; assign mathematical status and scoped verification records.
  \item Append the graph change and exploration event; validate identifiers, evidence, and dependencies.
  \item Reassess affected routes and generate the next frontier; if the mode permits new overall objectives, record their rationale and acceptance conditions before investigation.
  \item If the milestone appears satisfied, audit its full acceptance criteria.
  \end{enumerate}
\item Save the final state, actual stopping reason, completed work, and open frontier.
\end{enumerate}
\end{quote}

The controller specification accommodates all four RRE modes. A human can select and investigate targets, collaborators can share these activities, or an AI agent can conduct goal-directed exploration within a fixed charter. In the prospective open-ended mode, the agent would also formulate overall research objectives from the collection under the recorded policy. The operating mode and actor should be recorded for each action; a handoff preserves the current graph, open obligations, and remaining budget. Autonomous exploration may produce candidates and internal checks while still requiring human review before a stronger admission status is granted.

The case study illustrates goal-directed autonomous RRE. The author fixed the objective and authorized its acceptance criteria; the active AI research agent developed successive arguments, generated new targets, and used explicit recording utilities to publish graph revisions. Execution took place through an interactive agent session rather than an unattended controller service. Each saved revision has a separate relation to the recorded events and to the mathematical claims it supports.

\subsection{Significance, milestones, and stopping}\label{sec:significance}

We use provisional significance labels to support research management. They are relative to a declared research program and require a written rationale.

\begin{table}[htbp]
\centering\small
\begin{tabularx}{\linewidth}{@{}p{1cm}p{3cm}Y@{}}
\toprule
Level & Description & Evidence needed for an achieved result \\
\midrule
S0 & Connection or recovery & A useful relationship or known result is correctly identified. \\
S1 & Local advance & A stated subproblem is resolved in its required scope. \\
S2 & Reusable advance & A method or result supports a documented class of further arguments. \\
S3 & Major bottleneck closed & A predeclared major obstacle is resolved with the parameters needed downstream. \\
S4 & Objective resolved & A complete argument establishes the primary objective at the stated review level. \\
\bottomrule
\end{tabularx}
\caption{Significance is separate from proof status and literature novelty. A conjecture may have potential S3 significance without being an achieved S3 result.}
\end{table}

Significance is relative to the declared objective. Here the objective was to close the certificate-degree gap for an infinite family beyond flattenings, with a reusable method and matching bounds. The exact theorem in \cref{thm:main} and its transfer lemma meet those criteria. The result therefore closes the specified bottleneck within tensor-decomposition certificates. The designation ``level 3'' in the run charter expresses this research-management objective; the mathematical contribution is stated explicitly rather than inferred from the label.

A run stops when its declared terminal milestone passes the audit, its time or resource ceiling is reached, the user interrupts it, or execution cannot continue. An open-ended run may continue after an individual objective is resolved if its standing policy and remaining budget permit; ``open-ended'' does not mean unbounded execution. The reason for stopping is recorded literally. A ten-hour ceiling permits an earlier stop after acceptance; it does not imply ten hours of work were performed. Final audit time is included in the budget. Repeatedly restarting a subproblem must not silently reset the deadline.

\subsection{Replay and the limits of completeness}

RRE distinguishes three replay modes. \emph{Research playback} reconstructs recorded questions, decisions, evidence, and graph changes. \emph{Computational replay} reruns saved programs with pinned inputs and compares outputs under declared rules. \emph{Exploration rerun} starts another investigation from the original state and may follow a different path. The first two can be reproducible without the third being deterministic.

A useful event includes identifiers, timestamps, actor attribution, input graph revision, source hashes, target, action, concise rationale, outcome, evidence, and graph change. Model and tool versions, settings, random seeds, and returned candidate proposals should be retained when available. An implementation must report missing fields rather than suggest complete reproducibility from a hash chain alone.

Hashes detect changes relative to preserved digests. The record should also state which activities it captures. The case study saves mathematical targets, arguments, computations, decisions, six graph revisions, and their deltas. It records one concurrent event-log fork and its reconciliation without deleting either branch. These records support scientific playback and certificate replay; they do not purport to capture undocumented alternatives or internal model computation.

\section{A shared environment for research and learning}\label{sec:people}

\subsection{Human and AI contributions to one record}

A unified platform should allow researchers to use the same mathematical objects whether a contribution originates with a person, a model, a proof assistant, or a computation. A human researcher could enter a conjecture and ask for candidate routes. An AI system could identify a missing premise and request review of the proposed bridge. A collaborator could replace that bridge with a counterexample. The graph would preserve the rejected statement, the refutation, and the narrower question that follows.

Unified representation does not erase differences in responsibility. Contributions need separate attribution for problem formulation, argument drafting, implementation, checking, interpretation, and acceptance. A model's output should not acquire the author's review status simply because it appears in the author's workspace. Likewise, a human-written conjecture remains a conjecture. Mathematical admission depends on evidence rather than the identity or confidence of the proposer.

The actor performing a recorded action and the author of an imported artifact are separate attribution fields. The case-study events record the research actor explicitly, while source pins identify the exact imported artifacts. A file path, timestamp, or content hash establishes provenance rather than authorship. The record specification in \cref{app:records} keeps these fields separate.

Researchers should control whether objectives are supplied, jointly formulated, or delegated to an agent, together with admissible evidence, review requirements, and resource policies. One project may use AI only for retrieval; another may permit candidate generation and computational testing; a third may use the graph entirely through human contributions. These uses can share a data model without requiring agreement about the role of AI in mathematics. Cross-project imports should retain their original attribution and review conditions.

The persistent record also supports collaboration across time. A new researcher can inspect the remaining obligations rather than infer them from the final paragraph of a notebook. A returning researcher can see why a plausible approach was abandoned. A specialist can review a precisely scoped interface between two arguments without claiming responsibility for an entire research program. Whether this actually reduces collaboration or review costs remains an empirical question.

\subsection{Making research paths available to students}

Students often encounter polished proofs before they learn how to select a representation, identify a promising intermediate claim, or respond productively to a failed approach. Research exposition, seminars, and mentorship already address this gap. RRE proposes an additional resource: selected, evidence-linked exploration paths that can be revisited at different levels of detail.

Three kinds of record must remain distinct. A \emph{source-derived argument} reconstructs dependencies in a written proof. A \emph{retrospective research narrative} explains a possible or remembered route after the result is known. A \emph{contemporaneous exploration record} preserves questions and outcomes as the investigation proceeds. The first two can be educational, but neither establishes the actual historical path of discovery. Access to an expert's genuine research practice requires that expert's participation, consent, and annotation of the record.

A teaching interface could stop before a recorded decision and ask the student to propose the next target. It could reveal only the premises available at that time, then compare the student's proposal with the recorded continuation. It could also ask which hypothesis a counterexample violates, whether a finite experiment supports a universal claim, or how a quantitative bound affects a later theorem. Success would mean improved reasoning on new tasks, not simply the ability to repeat the displayed sequence.

The tensor-certificate example provides two concrete exercises. First, students can compare a tensor's exact rank with the ranks of its matrix flattenings, using the family $W_n$ in \cref{sec:boundary}. They can then explain why equations in tensor entries and certificates in decomposition parameters answer different questions. Second, students can start with the general balance reduction and calculate how contracting an $n$-mode target to $k$ modes changes multiplier degrees. The degree-9 certificate provides a concrete example, while \cref{cor:degree-transfer} states the reusable bound. The exercise identifies the cost of every contracted mode and explains why the normalized degree $\delta_R(T)/n$ obeys the transfer inequality. Both exercises connect a mathematical distinction to the choice of a new research target.

Educational use should include unsuccessful and unresolved branches in a curated form. Showing only the successful path can create the false impression that research decisions were inevitable. Showing every raw event can obscure the mathematics. The appropriate interface combines a readable narrative with access to the underlying evidence and labels any reconstruction. No student learning study has yet been conducted for RRE.

\subsection{Visualization as an interface to obligations}\label{sec:visualization}

Phase II introduced a multiscale viewer for collection relationships, argument maps, individual inferences, and source passages \cite{xie2}. RRE extends that design to research targets, time, and revisions. The earlier v2 companion established a before/after and contribution-view paradigm. The new certificate-study companion applies that paradigm to six immutable graph versions and to prerequisite closures for the exact theorem, the $W_n$ family, and the nilpotent-pencil branch. \Cref{sec:visual-effects} relates these views to the recorded exploration events.

A three-dimensional overview may help navigate a large collection, while a two-dimensional layout is often more readable for checking a selected inference. The coordinates are presentation choices. Neither geometric proximity nor an attractive layout establishes mathematical relevance. An inspectable view should expose stable identifiers, distinguish proposed from supported edges, show hypotheses even when cards are collapsed, and offer a textual trace for accessibility and audit.

The companion compares saved graph states, displays the jointly required premises of an inference, and opens the corresponding statement and evidence. The six revisions were published during the investigation. A revision may summarize several events; graph version and event time therefore remain separate fields. The printed figures combine a chronology of recorded developments with views of the actual graph, allowing readers to move between the history of a question and the argument that eventually answers it.

\subsection{Personal collections and future model training}

A researcher should be able to maintain a collection aligned with personal interests and accepted review standards, while importing selected records from shared collections. Federation requires stable identities, versioned imports, and explicit treatment of conflicting statements. Personal curation must not silently change the hypotheses of an imported theorem or imply that locally accepted claims have community approval.

With permission, curated records could also support training and evaluation of AI systems: premise selection, obligation generation, scope checking, error localization, or proposing a useful next target. Training examples would need to distinguish valid derivations from unsuccessful attempts and preserve the information available at the time of a decision. Evaluation splits should separate related proof families and chronological descendants to reduce leakage. Public availability of a notebook should not be treated as consent for every training use. This is a future use of the infrastructure, not an activity performed in the present study.

\section{Case study: exact certificate degree beyond flattenings}\label{sec:case}

\subsection{The objective and the resulting theorem}

The case study draws on a complexity research atlas developed during investigations motivated by $\VP_{\C}$ versus $\VNP_{\C}$ \cite{valiant}. Its selected question concerns the degree needed to certify that a tensor has no decomposition with a prescribed number of rank-one terms. The author fixed a standalone objective before the experiments: establish matching upper and lower certificate-degree bounds on an infinite family beyond ordinary flattening tests, using a reusable method.

Before exploration began, the author approved the five acceptance criteria in \cref{tab:case-milestone}. Together they define the case study's completion milestone. Intermediate targets could evolve, while these criteria remained fixed.

\begin{table}[htbp]
\centering\small
\begin{tabularx}{\linewidth}{@{}p{3.8cm}Y@{}}
\toprule
Criterion & Required deliverable \\
\midrule
1. Infinite family & A theorem beyond ordinary flattening tests, with explicit hypotheses and quantifiers in the original decomposition parameters. \\
2. Matching bounds & Matching lower and upper certificate-degree bounds, preferably exact. Matching asymptotics must provide a substantive advance in method and theorem. \\
3. Reusable method & A construction or lower-bound method with a proved extension or justified applicability class beyond one isolated tensor. \\
4. Literature comparison & A documented comparison with the closest tensor-rank and algebraic-proof-complexity results, distinguishing inherited ingredients from the new claims. \\
5. Replayable record & Full arguments, exact checks, source hashes, chronological decisions and failures, and before-and-after reasoning graphs. \\
\bottomrule
\end{tabularx}
\caption{The five prospectively approved criteria for the case study's completion milestone. Completion requires all five.}
\label{tab:case-milestone}
\end{table}

The charter permits a new family to be selected prospectively during exploration. The first family was a sequence of nilpotent matrix pencils; a later target generated from its smallest certificate led to the theorem below.

Let $T\in V_1\otimes\cdots\otimes V_n$, where the $V_i$ are nonzero finite-dimensional complex vector spaces. For a proposed decomposition with $R$ summands, introduce independent parameters $x_{r,i,j}$ and the coefficient equations
\begin{equation}\label{eq:raw-cp}
 F_J(x)=Q_J(x)-T_J,\qquad
 Q_J(x)=\sum_{r=1}^{R}\prod_{i=1}^{n}x_{r,i,j_i},
 \qquad J=(j_1,\ldots,j_n).
\end{equation}
These are the original CP-$R$ equations: CP denotes the decomposition as a sum of completely decomposable, or rank-one, tensors. Their common zero set consists of all such decompositions of $T$. If $\rank(T)>R$, this set is empty. A Nullstellensatz certificate is a polynomial identity
\begin{equation}\label{eq:unit-certificate}
 1=\sum_J h_J(x)F_J(x).
\end{equation}
Define $\delta_R(T)$ as the minimum of $\max_J\deg(h_JF_J)$ over all such identities, with total degree measured in the original parameters. Set $\delta_R(T)=\infty$ when the system has a solution. This convention charges all multiplier costs and specifies the representation in which the theorem holds.

\begin{theorem}[Exact degree for two-term decomposition]\label{thm:main}
Let $n\ge2$ and let $T\in V_1\otimes\cdots\otimes V_n$ have complex tensor rank greater than two. Then
\begin{equation}\label{eq:main-degree}
                       \delta_2(T)=3n.
\end{equation}
The conclusion holds for arbitrary finite mode dimensions, including dimension-one factors.
\end{theorem}

The theorem identifies the first degree at which the full polynomial certificate test can refute every inconsistent two-term decomposition. Its upper bound combines two mechanisms: a flattening minor when one is available, and a transferred boundary certificate otherwise. Its lower bound applies uniformly to the original equations. \Cref{app:math} gives the full argument and the explicit finite identity from which the boundary construction starts.

\subsection{Why the boundary case matters}\label{sec:boundary}

A flattening groups the tensor's modes into two nonempty parts and treats its coefficients as a matrix. A sum of $R$ rank-one tensors has every flattening rank at most $R$. Consequently, a nonzero $(R+1)\times(R+1)$ minor detects excessive rank. The atlas already contained a certificate of degree $n(R+1)$ obtained by telescoping such a minor, together with a lower functional proving that no smaller degree can work for any inconsistent CP-$R$ fiber.

The missing case is a tensor whose exact rank exceeds $R$ while its border rank is at most $R$. Border rank measures membership in the closure of the rank-$R$ tensors. All output polynomials vanishing on rank-$R$ tensors also vanish on that closure. They therefore cannot distinguish such a boundary tensor, whereas its decomposition equations still have no solution. The research target moves from an equation in tensor entries to a unit identity in decomposition parameters. The graph keeps these two objects and their different quantifiers explicit.

For the two-term model, an infinite example is
\begin{equation}\label{eq:wn}
 W_n=\sum_{i=1}^{n}e_0^{\otimes(i-1)}\otimes e_1
                  \otimes e_0^{\otimes(n-i)},\qquad n\ge3.
\end{equation}
Every bipartition $A\mid B$ gives
\[
 W_n=W_A\otimes e_0^{\otimes |B|}
       +e_0^{\otimes |A|}\otimes W_B,
\]
so every flattening has rank at most two. The degeneration
\[
 \frac{(e_0+t e_1)^{\otimes n}-e_0^{\otimes n}}{t}
       \longrightarrow W_n
\]
shows border rank two. Contracting to three active modes yields $W_3$, which has rank three. Thus $W_n$ has no two-term decomposition, and \cref{thm:main} gives the exact degree $3n$ even though all flattening tests at rank two vanish.

\subsection{Starting from the atlas}

At the start of the case study on October 31, the atlas contained 63,494 nodes and 22,183 AND inferences, combining the October 25 extraction with subsequent exploration components. The new run, RRE005, selected the distinct-label functional and flattening certificate from revision 241 of the preceding exploration component, RRE003, retained their source hashes, and rechecked their hypotheses. It began with eight records and no new inference. The initial graph specifies the objective, degree convention, two inherited tools, matrix-pencil baseline, and open boundary question. \Cref{app:complexity-graphs} describes the collection and distinguishes source extraction from subsequent theorem development.

The first chosen tensors have slices
\[
                  T_m=(I_m,J_m),\qquad m\ge2,
\]
where $J_m$ is the nilpotent Jordan block with superdiagonal entries one. Classical matrix-pencil theory gives exact rank $m+1$ and border rank $m$ \cite{buczynskilandsberg}. Here the atlas supplies the lower bound $3(m+1)$, but its flattening upper certificate cannot apply. A short invertible-slice argument verifies the obstruction: an $m$-term decomposition would make $J_m$ similar to a diagonal matrix. This fact establishes the test family; the new quantitative problem is the degree of its unit certificate.

\subsection{From small certificates to a uniform pencil argument}

The first generated task was to make the complete degree test computationally accessible. Each rank-one summand is unchanged by rescaling its mode vectors with scalars whose product is one. Averaging multipliers over these scalings retains only monomials whose degrees are equal across modes within each summand label. Summand permutations provide a further symmetry. This reduction preserves the existence of a certificate at the tested degree, so an unsuccessful complete test has its stated mathematical meaning. It does not discard terms merely because they appear unpromising.

Exact rational elimination found certificates of degrees 9 and 12 for $m=2,3$. Raw-variable expansion checked their identities independently of the orbit multiplication routine. These outcomes generated the sharp-bound target $\delta_m(I_m,J_m)=3m+3$ and a task to exploit the target's additional grading. A modular degree-15 success for $m=4$ was then lifted to a rational certificate and expanded exactly. The three finite values reach the inherited lower bound.

The next generated target sought a uniform proof. For a nonzero nilpotent matrix $N$, choose $H$ with $NH-HN=N$. The decomposition equations and polynomial adjugates produce a commutator identity without dividing by a parameter. Tracking every multiplication yields the following separate result.

\begin{theorem}[Nilpotent-pencil certificate bounds]\label{thm:pencil}
For every $m\ge2$ and every nonzero nilpotent $N\in\C^{m\times m}$,
\begin{equation}\label{eq:pencil-bound}
                3m+3\le\delta_m(I_m,N)\le3m+6.
\end{equation}
The degree is a multiple of three, so it is one of the two displayed endpoints.
\end{theorem}

\Cref{app:pencil} supplies the constructive proof and its degree ledger. This result narrowed an unbounded degree search to one remaining balanced degree block. The graph then generated the exact-endpoint question and retained multiple-block tests with their rational or modular evidence types. For Jordan type $(2,1)$, exact rational elimination places $1$ in the degree-12 row span; type $(2,2)$ passes the degree-15 test over $\mathbb F_{1000003}$. The all-$m$ exact endpoint remains open; that branch retains its own statement when the later two-term theorem is added.

\subsection{Charging contracted modes in a transfer}

The search used a balanced basis from the start: \cref{lem:balance} guarantees that every certificate can be made label-balanced without increasing its degree. The degree-9 identity supplies a concrete base certificate. The useful transfer question is quantitative: if a higher-order tensor contracts to its target, how much degree is needed to pull the identity back in the original parameters? Target T011 records this proposal before the transfer checks and the all-target theorem were saved. The key step is to charge all contracted modes through the existing balance, so a degree block of size $k$ becomes one of size $n$.

The answer is a general lemma. Suppose a $k$-mode CP-$R$ certificate has label-balanced multipliers of degree at most $k(q-1)$. If an $n$-mode target maps to its target by linear maps on $k$ retained modes and linear functionals on the other modes, then the certificate pulls back with product degree at most $nq$ in the original parameters. Contracted parameter factors are absorbed into the first retained mode of each summand. Balance makes a multiplier of degree $kt$ become one of degree at most $nt$; the substituted equation is a constant linear combination of original equations. This is the precise cost that the graph's degree convention requires.

Combining balance reduction with the transfer lemma removes any separate balance assumption on the certificate. If $T$ has order $n$, $S$ has order $k\le n$, and $T$ maps to $S$ by the stated mode maps and contractions, then
\[
 \delta_R(T)\le\frac{n}{k}\,\delta_R(S),\qquad \delta_R(S)<\infty.
\]
\Cref{cor:degree-transfer} proves this general bound by applying transfer to a minimum-degree certificate. It makes the method reusable for every finite CP-$R$ certificate under the specified maps.

The smallest pencil $(I_2,J_2)$ becomes $W_3$ after a coordinate swap. Its degree-9 certificate therefore transfers to $W_n$ with degree $3n$. The inherited lower bound is also $3n$, establishing an exact infinite-family result. The same lemma can now be reused wherever a target has the required contraction. It is the new bridge between a finite polynomial identity and an all-order theorem.

\subsection{From the family to the full two-term model}

The next step combines the transfer with classical geometry. Landsberg and Manivel show that the $3\times3$ minors of all bipartition flattenings define the second Segre secant variety set-theoretically; Raicu proves the stronger ideal-generation result \cite{landsbergmanivel,raicu}. Within this secant variety, a tensor that is not an actual two-term sum lies on a tangent space to the Segre variety \cite[arXiv version, Corollary~1 and Notation~1]{ballicobernardi}. Such a tensor can be written
\begin{equation}\label{eq:tangent-main}
 T=\lambda\bigotimes_i v_i+
       \sum_i u_i\otimes\bigotimes_{j\ne i}v_j,
\end{equation}
with the factors placed in their corresponding modes. Rank greater than two requires at least three active modes, in which $u_i$ is independent of $v_i$. Retaining three of them, contracting the others, and removing the pure term by a constant shear gives $W_3$.

This yields an exhaustive case split. If some flattening has rank greater than two, the minor certificate has degree $3n$. Otherwise, the tangent projection and balanced transfer have degree $3n$. Combining either upper route with the inherited lower bound proves \cref{thm:main}. In the reasoning graph, the two routes are conditional derivations followed by an exhaustive-case inference. They are not two hypotheses that the target must satisfy simultaneously.

The $W_n$ corollary has a shorter prerequisite path: the explicit base identity, the transfer lemma, and the lower functional suffice. The full theorem additionally uses the classical geometric classification. Separate traces expose exactly where those imports enter. The contribution is the original-parameter certificate-degree characterization and its transfer method; the tensor-rank and secant classifications retain their established attribution.

\subsection{Verification and the quantitative record}\label{sec:case-checks}

The proof is accompanied by exact checks at three levels. First, the certificates for the single-block pencils $(I_m,J_m)$, $m=2,3,4$, expand to the constant polynomial one over $\Q$. The largest has 6,211 orbit terms and 429,076 expanded product terms. Second, the commutator identities were checked symbolically for $m=2$ and at 112 exact substitutions across 14 Jordan-block families through $m=8$, including singular parameter matrices. Third, the transferred certificate was expanded in the original $W_n$ variables for 32 orders, $n=3,\ldots,32,64,128$. All identities equal one and have degree $3n$. A further 288 exact coefficient-map identities test contraction, nontrivial basis changes, and removal of the pure tangent term across 36 cases.

These computations test the explicit certificates and their implementations; the uniform conclusions follow from the proofs in \cref{app:math}. A modular success is retained as a modular result until a rational certificate is supplied. Mathematical arguments, exact identities, classical imports, and structural validation have separate records. The study includes internal mathematical review, exact replay, and an initial audit of the mathematical arguments by the author. Its literature comparison concerns the specific certificate-degree statement and the primary works described above.

The closest existing formulations clarify the advance. Tensor-rank Nullstellensatz systems were already available \cite{aliabadifriedland}; the present theorem gives a sharp threshold for their original two-term equations. The flattening equations describe a closure in tensor space \cite{landsbergmanivel,raicu}; the boundary certificate instead proves emptiness of an exact decomposition fiber. The secant--tangent dichotomy supplies the exhaustive cases \cite[arXiv version, Corollary~1]{ballicobernardi}; the balanced lemma supplies the parameter-degree bound. These are distinct contributions in the dependency graph.

\subsection{Graph evolution, replay, and visualization}\label{sec:visual-effects}

Six immutable versions record the investigation. The graph grows from eight records to 68 nodes and 34 AND inferences. Each publication retains a delta and source hashes and advances the new component in the atlas. The final version records acceptance against the original infinite-family objective. A 44-record event history preserves target proposals, computations, proofs, corrections, publication events, and the stopping decision. \Cref{app:records} explains the event identifiers and replay procedure.

\begin{table}[htbp]
\centering\small
\begin{tabularx}{\linewidth}{@{}p{1.4cm}r r Y@{}}
\toprule
Revision & Nodes & AND steps & Mathematical or historical effect \\
\midrule
v001 & 8 & 0 & Fix the objective, raw model, inherited bounds, and boundary family. \\
v002 & 19 & 7 & Establish complete symmetry reduction and exact degrees 9 and 12; generate a uniform target. \\
v003 & 43 & 22 & Certify degree 15 and prove the all-$m$ pencil bound; generate the endpoint question. \\
v004 & 46 & 22 & Add multiple-block screens and retain the remaining gap. \\
v005 & 66 & 33 & Prove balanced transfer, the $W_n$ formula, and the all-target two-term theorem. \\
v006 & 68 & 34 & Complete the mathematical audit, replay delivery, and milestone assessment. \\
\bottomrule
\end{tabularx}
\caption{Saved graph states during the exploration. Counts include scopes, evidence, objectives, and open questions as well as derived claims.}
\label{tab:versions}
\end{table}

\begin{figure}[p]
\centering
\begin{tikzpicture}[>=Latex,font=\small,
 card/.style={draw=blue!40!black,rounded corners=3pt,fill=blue!3,align=left,text width=5.65cm,minimum height=2.3cm,inner sep=9pt},
 arr/.style={->,thick,draw=blue!55!black}]
\node[card] (a) at (0,0) {\textbf{1. A gap in inherited reasoning}\\[3pt]
Lower degree $n(R+1)$; the matching upper certificate needs a large flattening.\\[3pt]\footnotesize v001: 8 nodes, 0 new inferences};
\node[card] (b) at (7.1,0) {\textbf{2. Boundary certificates}\\[3pt]
Symmetry makes complete tests feasible. Exact degrees 9 and 12 suggest a uniform target.\\[3pt]\footnotesize v002: 19 nodes, 7 inferences};
\node[card] (c) at (0,-3.4) {\textbf{3. A uniform pencil theorem}\\[3pt]
Adjugates and a commutator give\\$3m+3\le\delta_m\le3m+6$.\\[3pt]\footnotesize v003: 43 nodes, 22 inferences};
\node[card] (d) at (7.1,-3.4) {\textbf{4. Retain the remaining question}\\[3pt]
Multiple-block tests preserve the exact endpoint as open. The base identity remains reusable.\\[3pt]\footnotesize v004: 46 nodes, 22 inferences};
\node[card,fill=green!5] (e) at (0,-6.8) {\textbf{5. Charge contracted modes}\\[3pt]
Transfer costs the factor $n/k$. The degree-9 base and geometry give $\delta_2(T)=3n$.\\[3pt]\footnotesize v005: 66 nodes, 33 inferences};
\node[card,fill=green!5] (f) at (7.1,-6.8) {\textbf{6. Publish and preserve}\\[3pt]
Audit the original criteria; save proofs, checks, graph paths, and further targets.\\[3pt]\footnotesize v006: 68 nodes, 34 inferences};
\draw[arr] (a)--(b);
\draw[arr] (b.south)--($(b.south)!0.5!(d.north)$)-| (c.north);
\draw[arr] (c)--(d);
\draw[arr] (d.south)--($(d.south)!0.5!(f.north)$)-| (e.north);
\draw[arr] (e)--(f);
\end{tikzpicture}
\caption{Recorded research progression. Each stage changes what the next investigation can use. Arrows describe exploration history, while the theorem's implication structure is shown separately in \cref{fig:certificate-proof}. The open pencil endpoint survives the success of the two-term branch.}
\label{fig:discovery-timeline}
\end{figure}
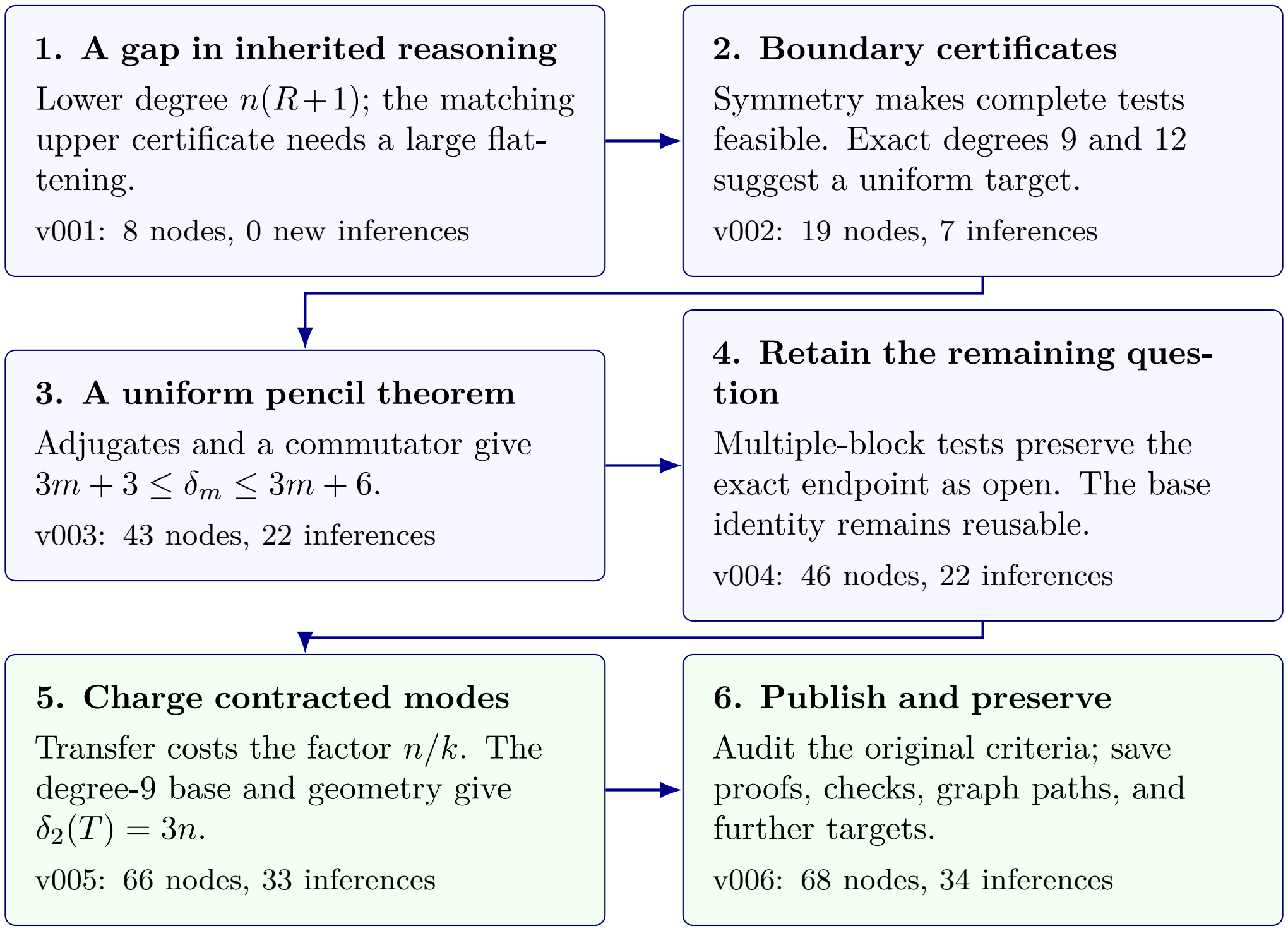

\begin{figure}[p]
\centering
\includegraphics[width=\linewidth]{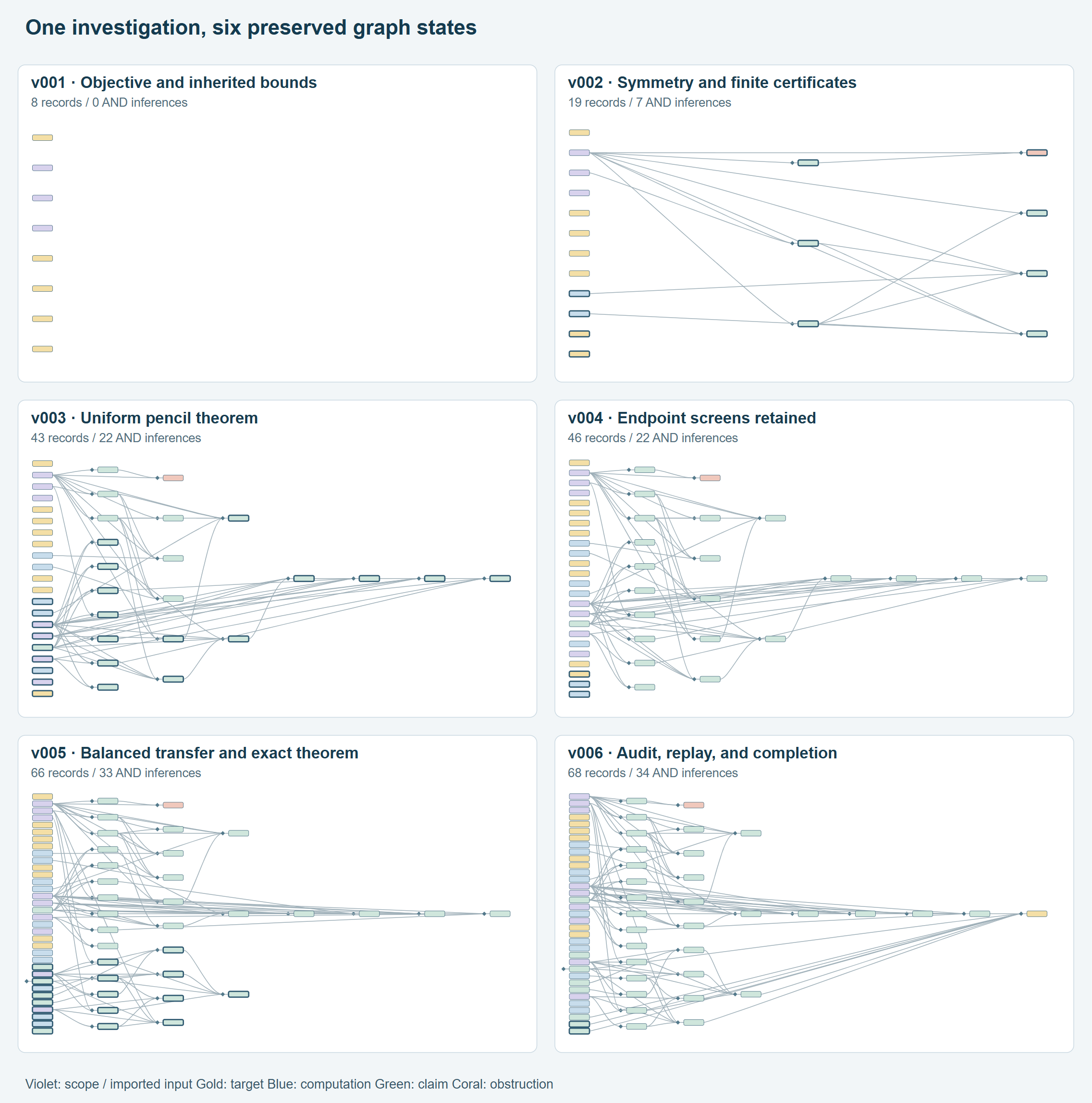}
\caption{Incremental growth of the certificate-study graph, rendered from the six saved graph files. Each small map contains all records and inference segments in that version; positions follow proof depth, and scopes, targets, computations, and claims retain different colors. After v001, heavier outlines mark newly added records. Counts measure records, not independent theorems. The compact maps show structure; the companion opens each version at full resolution with statements and evidence.}
\label{fig:certificate-versions}
\end{figure}

\begin{figure}[p]
\centering
\begin{tikzpicture}[>=Latex,font=\small,
 box/.style={draw=blue!40!black,rounded corners=2pt,align=center,text width=5.55cm,minimum height=1.45cm,inner sep=8pt,fill=blue!3},
 result/.style={box,fill=green!6,draw=green!40!black},
 arr/.style={->,draw=blue!55!black,thick}]
\node[box,text width=12.5cm,fill=violet!5] (scope) at (3.5,0)
 {\textbf{Scope:} order $n\ge2$, tensor rank $>2$, over $\C$\\Original two-term equations; total multiplier--equation degree};
\node[box] (split) at (3.5,-2.2) {\textbf{Exhaustive cases}\\Some flattening rank $>2$, or every flattening rank $\le2$};
\node[box] (minor) at (0,-4.6) {\textbf{Case A: inherited minor lemma}\\A nonzero $3\times3$ minor\\gives an upper certificate of degree $3n$};
\node[box] (geom) at (7,-4.6) {\textbf{Case B: classical geometry}\\Secant-two flattening theorem\\and tangent-space description};
\node[result] (transfer) at (7,-7.0) {\textbf{Projection and balanced transfer}\\Map the target to $W_3$; pull back\\the explicit degree-9 certificate};
\node[result] (upper) at (3.5,-9.3) {\textbf{Upper bound in every case}\\$\delta_2(T)\le3n$};
\node[box] (lower) at (0,-7.0) {\textbf{Inherited lower functional}\\Annihilates $h_JF_J$ when\\$\deg(h_JF_J)<3n$, but $L(1)=1$};
\node[result,text width=8cm] (exact) at (3.5,-11.7) {\textbf{Exact theorem}\\$\delta_2(T)=3n$ for every $\rank(T)>2$};
\draw[arr] (scope)--(split);
\draw[arr] (split.west)-|(minor.north);
\draw[arr] (split.east)-|(geom.north);
\draw[arr] (geom)--(transfer);
\draw[arr] (minor.west)--++(-.45,0)|-(upper.west);
\draw[arr] (transfer.south)|-(upper.east);
\draw[arr] (upper)--(exact);
\draw[arr] (lower.south)--++(0,-.7)--++(-3.3,0)|-(exact.west);
\end{tikzpicture}
\caption{Proof outline of the exact theorem, with the original-parameter scope attached throughout. Blue boxes denote inherited or classical inputs; green boxes show the certificate transfer and resulting bounds. The two upper routes are conditional arguments combined by exhaustive cases; for $n=2$, only Case A occurs. The lower functional and either applicable upper route yield equality. Each box summarizes the indicated lemma or argument in \cref{app:math}; the companion expands its recorded premises.}
\label{fig:certificate-proof}
\end{figure}
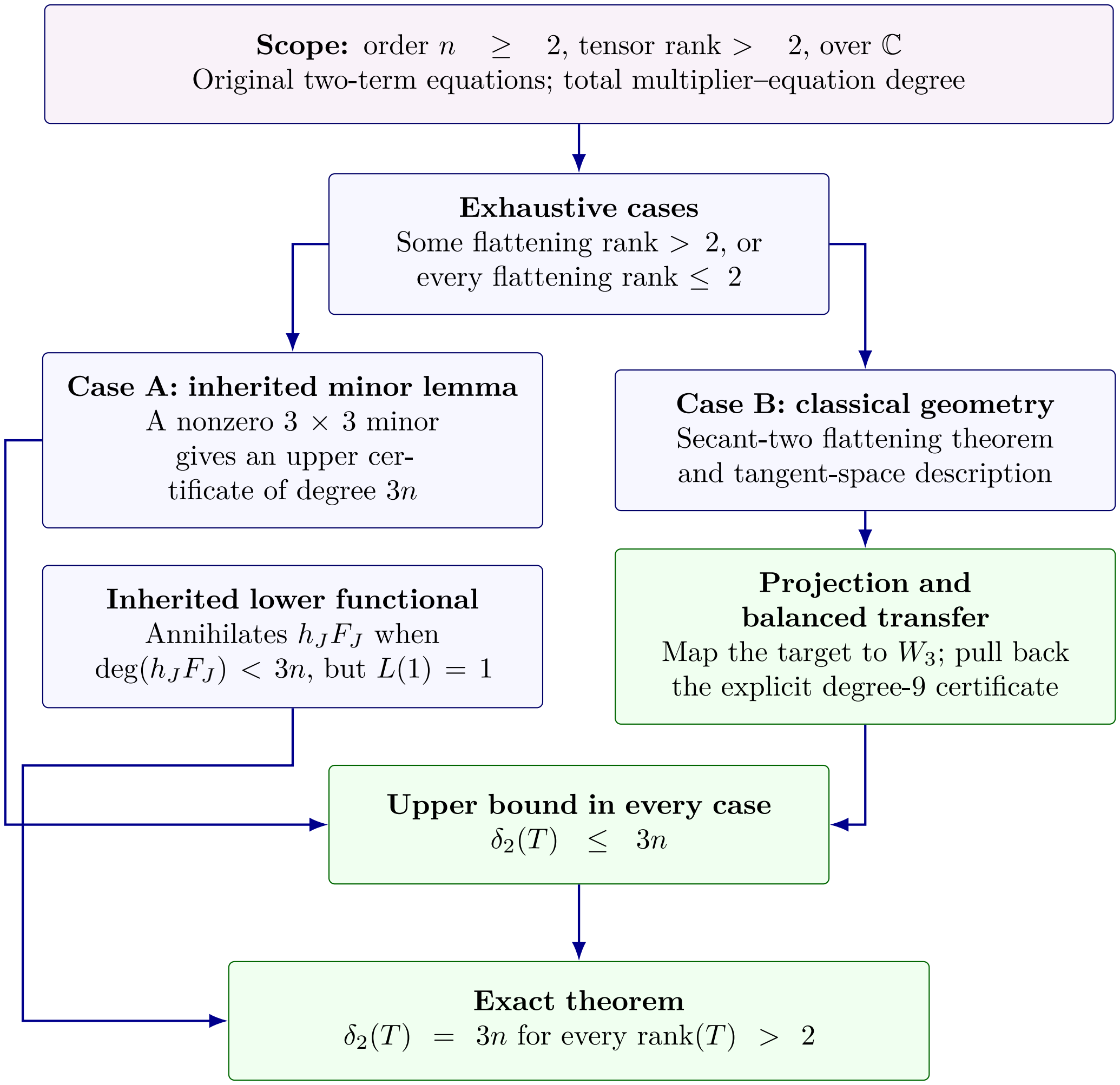

\begin{figure}[p]
\centering
\includegraphics[width=\linewidth,height=.75\textheight,keepaspectratio]{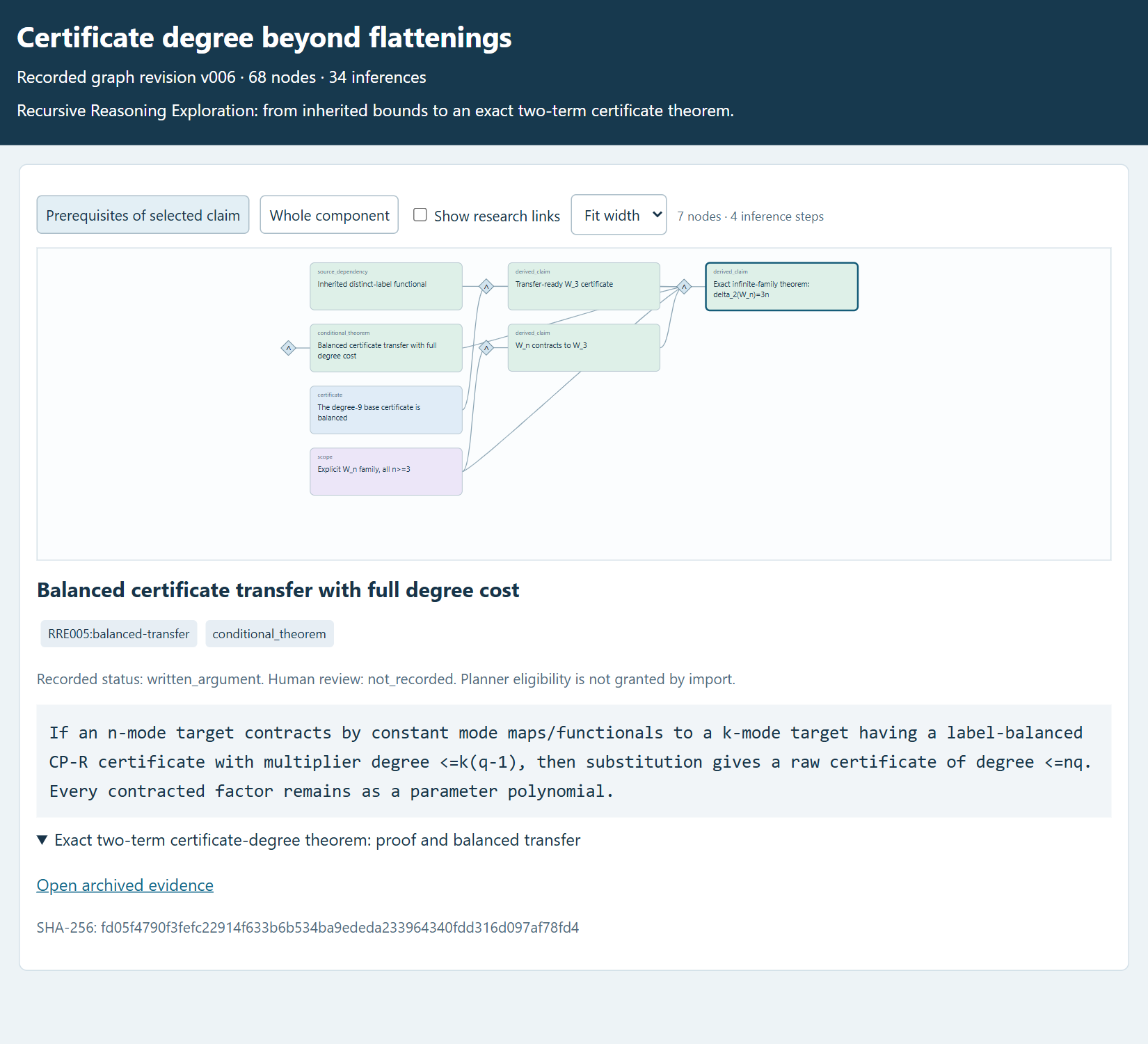}
\caption{The implemented companion connects a selected graph record to its statement, scope, and hash-pinned evidence. The view focuses on the balanced transfer used to extend the boundary certificate. It is an interface excerpt; the versioned companion provides the full graph and prerequisite traces.}
\label{fig:certificate-inspector}
\end{figure}

The companion provides a view for every saved version and prerequisite traces for each result. The six-version overview in \cref{fig:certificate-versions} shows how the argument grows from the inherited input. \Cref{fig:certificate-proof} summarizes the proof routes and their scope, while \cref{fig:certificate-inspector} shows the evidence inspector. The companion exposes the individual AND steps and attached hypotheses. The timeline in \cref{fig:discovery-timeline} instead shows how mathematical outcomes generated subsequent targets. Distinguishing these views lets a reader reconstruct both the final proof and the recorded process that produced it.

Two dedicated exports retain the nilpotent-pencil and exact two-term theorem paths with their evidence. They can be consumed by a separate visualization process while exploration continues elsewhere in the atlas. The final component is integrated, while the earlier working graphs and their revisions remain intact.

\subsection{What the case contributes to the RRE method}

The case has several recursive transitions with explicit mathematical content. Finite certificates generate a uniform theorem target; its additive gap generates an endpoint question; the question of charging contracted modes generates a transfer target; and the transfer combines with classical geometry to cover every inconsistent two-term fiber. The record retains why each question became useful and which premises made the resulting answer valid.

The decisive change of viewpoint is also preserved. A tensor in a rank closure cannot be excluded by an output equation, yet its empty exact decomposition fiber admits a polynomial certificate. Recording this distinction avoids repeatedly pursuing the wrong type of obstruction. Recording the base identity, the general balance reduction, and the contraction degree calculation makes the later reuse inspectable. The theorem and the reusable bridge supply the substantive mathematical outcome; the evolving graph supplies an inspectable account of their dependencies and development. 

The final results meet the criteria in \cref{tab:case-milestone} as follows. The exact formula $\delta_2(W_n)=3n$ for every $n\ge3$ supplies an infinite family beyond flattenings with matching bounds (criteria 1 and 2); \cref{thm:main} extends this to the full two-term model. The general transfer lemma supplies the reusable method (criterion 3). The comparison in \cref{sec:case-checks} identifies the contribution relative to prior work (criterion 4). The proofs, exact checks, pinned sources, event history, and six graph revisions provide the replayable record (criterion 5). These deliverables support the completion decision recorded in the final revision. The comparative value of this organization is the subject of the evaluations proposed next.

\section{Evaluation: what would establish broader value?}\label{sec:evaluation}

RRE needs evaluation at the level of mathematical progress and human use. More generated nodes, longer traces, or persuasive narratives are weak proxies. The framework should be judged by whether it helps produce correct, relevant, reusable results and whether others can understand and extend them at reasonable cost.

\subsection{Prospective research comparisons}

A first study of goal-directed exploration should use several predeclared mathematical objectives with explicit intermediate milestones. Problems should vary in their dependence on source retrieval, quantitative bounds, and transfer between settings. Include tasks on which a plausible route is false or a milestone is out of reach within the budget. Report all assigned tasks and stopped runs, including negative outcomes.

At least three conditions would be informative: assistance using ordinary research notes and document retrieval; assistance using a static curated reasoning graph; and RRE using the same initial graph with target generation and incremental updates. Match available sources, model versions, computational resources, time, and access to verification tools. Record the operating mode, human contributions, and any handoffs; goal-directed autonomous and collaborative RRE runs need not have the same labor requirements. Count the cost of constructing and curating the initial graph. Otherwise, an apparent improvement could reflect additional preparation rather than the recursive method.

Assessment should use reviewers with relevant expertise and, where possible, conceal the experimental condition. They should judge correctness, scope, novelty relative to documented prior work, and usefulness to the declared objective. Report review time, disagreements, and uncertainty. A result obtained only after substantial human repair should include that repair in both the outcome and resource accounting.

Useful endpoints include time to the first accepted intermediate result, fraction of proposed routes rejected for scope errors, number of previously documented failures repeated, and success of a later researcher using the recorded result. These measure different things and should not be combined into a single score without justification. The number of attempted subproblems is informative only with their difficulty and outcomes.

Open-ended exploration needs a separate comparison with matched starting collections and budgets. Reviewers should assess the mathematical value of the generated objectives as well as the correctness and reuse of their outcomes. Record rejected objectives and unsuccessful investigations; allowing an agent to choose easy questions must not be mistaken for improved discovery. Research-strategy improvement requires comparisons before and after a method change on held-out tasks, with errors, resource use, and human repair included. Neither capability is evaluated in the present case study.

\subsection{Ablations and cumulative effects}

To test the value of recursion, freeze the graph in one condition and allow reviewed increments in another. To test the value of negative knowledge, remove access to prior failed routes while retaining successful arguments. To test scope tracking, compare full records with otherwise identical records lacking explicit parameter and quantifier fields. Such ablations should be used in evaluation copies, not by corrupting an active research collection.

A cumulative knowledge claim requires a second stage. Freeze the outputs of one investigation, then ask a new researcher or a fresh system to solve a related target using that record. Measure whether the added lemma, counterexample, or quantitative bound is correctly retrieved and reused. The case study provides several connected developments within one investigation: finite certificates, a uniform pencil bound, balanced transfer, and the all-target two-term theorem. A subsequent-user study would test whether those additions also help a separate investigation.

\subsection{Review, replay, and correction experiments}

A replay study should give an independent evaluator the artifact package and a declared task: reconstruct the accepted argument, reproduce a certificate, identify a missing premise, or explain the stopping decision. Success requires more than opening the files. The evaluator must be able to locate the necessary source versions and distinguish what was proved from what was tested.

Correction propagation can be evaluated with controlled changes to a copied graph. For example, withdraw a premise used by two derivations while retaining an unaffected alternative proof. Check whether the system flags exactly the affected routes and preserves supported alternatives. Deliberately introduced scope errors can test whether uniform and fixed-parameter claims are distinguished. Automated structural checks and expert mathematical judgments should be reported separately.

\subsection{Human learning and collaboration}

Educational evaluation should compare a polished proof, the same proof with a static dependency graph, and a curated exploration history. Outcomes should include transfer to a new problem, identification of missing hypotheses, quality of proposed intermediate targets, and retention after a delay. Time on task and prior knowledge matter. A visually engaging display is not evidence of improved mathematical understanding.

Collaboration studies should ask whether contributors can resume one another's work, assign review to specialists, and preserve attribution under concurrent revisions. They should also measure the burden of maintaining the record. If recording every action consumes more time than it saves, RRE will need better selective capture and summarization. A useful system should support different levels of detail while making the recording policy explicit.

\begin{table}[htbp]
\centering\small
\begin{tabularx}{\linewidth}{@{}p{3cm}YY@{}}
\toprule
Claim to evaluate & Evidence available here & Further evidence needed \\
\midrule
Recorded recursive investigation is feasible & One run with successive generated targets and six graph revisions & Multiple independently executed runs \\
Mathematical outcomes are correct & Full theorem proofs, an explicit certificate, and exact replay & Independent mathematical review; selected formalization \\
Graphs improve discovery & No controlled comparison & Matched baselines and preregistered milestones \\
Records support cumulative reuse & Reuse within the selected run & Later users and held-out objectives \\
Replay reduces review effort & Saved artifacts and internal replay & Independent reconstruction and cost measurement \\
Exploration paths support learning & Proposed exercises and interfaces & Student studies with transfer outcomes \\
\bottomrule
\end{tabularx}
\caption{Current evidence and the evaluations required for stronger claims.}
\end{table}

\section{Limitations and research agenda}\label{sec:discussion}

\subsection{Annotation and trust}

An incomplete reasoning graph can mislead precisely because it looks explicit. Missing premises, merged quantifiers, and misclassified evidence can hide behind a clean diagram. Phase II's review experience motivates source anchors and explicit scope, but those features do not establish annotation accuracy automatically. Automatic extraction should therefore produce reviewable drafts, with coverage accounting and targeted human checking. 

The case study combines a selected working graph, an agent-led investigation, written proofs, and exact certificate checks. Its strongest mathematical outcome is the uniform theorem in \cref{thm:main}. The methodological question is broader: whether maintaining this structured history improves later research, review, and learning. That question calls for the comparative and cumulative-use studies in \cref{sec:evaluation}.

\subsection{Scale, selectivity, and failure modes}

A collection can become unusable if every generated suggestion is retained as an undifferentiated node. RRE needs separation between an archival event store and a curated current view. Unreviewed proposals can remain searchable without appearing as supported mathematical facts. Duplicate detection, summary views, and relevance filters should reduce navigation cost while preserving access to the underlying evidence.

Search can also become trapped by the existing graph. If a promising route lies outside the represented literature, repeated local recombination may never find it. The protocol should permit source expansion, alternative representations, and human redirection. Any claim that graph-guided exploration is complete would require assumptions far beyond the present proposal.

The method can create incentives to optimize visible activity rather than mathematics. Provisional significance labels should therefore be justified by resolved obligations and downstream uses, not by the size of the graph or the length of a run. A precise counterexample may be more valuable than many speculative connections. A clear statement that no progress was made is a legitimate outcome.

\subsection{Community concerns and participation}

RRE addresses a specific set of concerns: hidden dependencies, loss of context, duplicated failed work, unclear attribution, and the difficulty of inspecting AI-assisted claims. It does not settle disagreements about commercial AI, environmental costs, labor, or whether particular forms of automation belong in mathematical practice. An inspectable record is one contribution to responsible research, not a reason to dismiss those concerns.

Participation should be voluntary. Researchers should be able to use the infrastructure without AI-generated contributions, restrict automated activity, and decide which records become public. Shared schemas and portable exports can reduce dependence on one provider. Human mathematical judgment remains essential in setting worthwhile objectives, interpreting significance, and deciding what deserves community attention.

\subsection{Next steps}

The immediate technical agenda is to implement a controller that enforces the protocol and supports concurrent contributions, while extending the versioned companion with persistent target management and automatic correction propagation. It should integrate symbolic computation and proof assistants where useful, retain explicit limits when formalization is incomplete, and export both a readable narrative and machine-readable evidence. The proposed schema in \cref{app:records} is a starting point for that implementation rather than a completed interoperability standard.

Open-ended autonomous exploration would extend this controller to selecting overall research objectives from the collection. Research-strategy improvement is a distinct extension: the system could propose changes to retrieval, target selection, conjecture generation, or checking tools, then evaluate those changes before adopting them. Each revision should retain the previous method, evaluation tasks, resource costs, results, and acceptance decision. These records concern the research procedure and belong alongside the mathematical graph; they are not premises proving its theorems. More graph records alone would establish neither better research methods nor greater mathematical value.

For tensor certificates, the next mathematical targets are precise. The nilpotent-pencil argument leaves two possible degrees, $3m+3$ and $3m+6$, for arbitrary nonzero nilpotent types. For three or more summands, the transfer lemma already applies; the next task is to find reductions to base tensors with certificates sharp enough to determine exact degrees. These targets can use the recorded identities and degree ledger directly. For RRE, independent reuse of this package and controlled comparisons provide the next methodological tests.

\section{Conclusion}

The Math Instruments program progresses from identifying reusable devices, through reconstructing their reasoning, to investigating new questions with an evolving mathematical record. RRE proposes a disciplined form of that third phase. Each cycle connects an objective to a target, an investigation to evidence, and an outcome to the next frontier. Supported results, scoped failures, and unresolved obligations all contribute to the collection.

The case study turns an inherited lower bound and a missing boundary argument into an exact theorem for the entire two-term tensor-decomposition model. The reusable transfer rule comes from combining the general balance reduction with explicit degree accounting for contracted modes. Applying it to the degree-9 boundary certificate and then using classical geometry resolves the original infinite-family objective and more. The graph preserves the intermediate pencil theorem, the route to the general result, and the remaining questions. This is the proposed form of cumulative research: mathematical outcomes return to a record that others can inspect, extend, and use to formulate their next investigation.

\section*{Data and software availability}
The local companion package contains the certificate-study charter, 44 exploration records, six graph versions and their deltas, mathematical notes, exact certificates, checking programs, source pins, and validation reports. It also contains the versioned graph viewer and the figure-generation and capture programs. \Cref{app:records} describes replay; \cref{app:complexity-graphs} documents the underlying atlas and its representative sample. The two earlier phases of the Math Instruments program are available through the versioned Zenodo records in \cite{xie1,xie2}. The new package is a local preprint companion; its manifest records the included artifacts and their hashes.

\section*{Acknowledgment}
The author thanks OpenAI's GPT-6.0 Astra for conducting the goal-directed autonomous Recursive Reasoning Exploration (RRE) run under the author's research objective and authorized milestone, including developing the arguments, writing checking programs, generating the follow-up target, and assisting with manuscript drafting. GPT-6.0 Astra was also used to develop the v2 companion visualization of reasoning-graph dynamics. The author established the research direction, designed the RRE methodology, set the case study's objective and authorized milestone, performed an initial audit of the mathematical arguments for correctness, and takes responsibility for the final presentation of the paper.

\appendix

\section{Certificate-degree proofs}\label{app:math}

This appendix proves the mathematical statements used in the case study. All polynomial rings are over $\C$ unless rational coefficients are specified. Tensor rank is the minimum number of completely decomposable summands. The equations and degree convention are those of \cref{eq:raw-cp,eq:unit-certificate}; the order $n$ and proposed number of summands $R$ are separate parameters.

\subsection{The inherited universal lower bound}

\begin{lemma}[Distinct-label functional]\label{lem:lower}
For every order-$n$ tensor $T$ and integer $R\ge1$, the original CP-$R$ system satisfies
\[
                         \delta_R(T)\ge n(R+1).
\]
The statement is independent of the finite mode dimensions. When $\rank(T)\le R$, the inequality holds with $\delta_R(T)=\infty$ by the convention in \cref{sec:case}.
\end{lemma}

\begin{proof}
On monomials of total degree at most $n(R+1)-1$, define a linear functional $L$ as follows. A monomial is admissible if, at every occupied summand label $r$, it contains exactly one variable from each of the $n$ modes. Thus each occupied label determines an output multi-index $J_r$. For $t$ occupied labels, set
\[
 L\!\left(\prod_{r\text{ occupied}}\prod_{i=1}^n x_{r,i,j_{r,i}}\right)
    =\frac{\prod_{r\text{ occupied}}T_{J_r}}{(R)_t},
 \qquad (R)_t=R(R-1)\cdots(R-t+1).
\]
Set $L(1)=1$ and assign zero to every nonadmissible monomial. There are at most $R$ occupied labels, so all denominators used are nonzero.

Let $M$ be any multiplier monomial of degree at most $nR-1$. If $M$ is admissible, it occupies $t\le R-1$ labels. In the product $MQ_J$, the $t$ summands using an already occupied label are nonadmissible. Each of the remaining $R-t$ labels contributes
\[
 \frac{T_J\prod_{r\text{ occupied}}T_{J_r}}{(R)_{t+1}}.
\]
Their sum is $T_JL(M)$. If $M$ is nonadmissible, multiplication by one variable from every mode at a single label cannot make it admissible: a partially occupied label acquires a repeated mode, an already repeated mode persists, and any other bad label is unchanged. Hence both $L(MQ_J)$ and $T_JL(M)$ are zero. Thus $L(MF_J)=0$ for every permitted $M$ and $J$.

Every nonzero $F_J$ has degree $n$, so a certificate of product degree at most $n(R+1)-1$ would have multipliers of degree at most $nR-1$. Applying $L$ to that identity would give $1=0$. This proves the bound.
\end{proof}

This argument is the lower functional inherited from the atlas. It is written here in full because its independence from mode dimensions is needed for both the pencil family and contractions with passive modes.

\subsection{The flattening certificate and balanced multipliers}

\begin{lemma}[Minor certificate]\label{lem:minor}
If a bipartition flattening of $T$ has a nonzero $(R+1)\times(R+1)$ minor, then $\delta_R(T)\le n(R+1)$.
\end{lemma}

\begin{proof}
Write $q=R+1$ and select the corresponding submatrices $Q_0(x)$ and $T_0$. The flattening of $Q(x)$ is a sum of $R$ rank-one matrices, so $\det Q_0(x)$ is identically zero. Replace the rows of $T_0$ by rows of $Q_0$ one at a time. Multilinearity of the determinant gives
\[
 -\det T_0=\det Q_0-\det T_0=\sum_J c_J(x)F_J(x),
\]
where each cofactor $c_J$ uses at most $R$ entries of $Q$ and remaining constant entries of $T$. Therefore $\deg c_J\le nR$. Dividing by the nonzero constant $-\det T_0$ gives the required certificate.
\end{proof}

\begin{lemma}[Balance reduction]\label{lem:balance}
If an original CP-$R$ system admits a unit certificate of degree at most $D$, it admits one at the same degree whose multiplier monomials have equal degrees in all modes separately at each summand label. The minimum finite certificate degree is consequently a multiple of $n$.
\end{lemma}

\begin{proof}
For each label $r$, act on its mode vectors by scalars $\lambda_{r,1},\ldots,\lambda_{r,n}$ with $\prod_i\lambda_{r,i}=1$. Every $F_J$ and the constant one are invariant. Projecting a polynomial identity to the zero character of this torus projects each multiplier to its invariant monomials and does not increase degree. A monomial is invariant precisely when its mode degrees agree within each label. Such a monomial has degree $n\sum_r t_r$ for nonnegative integers $t_r$. A nonzero balanced multiplier times $F_J$ has degree $n(1+\max\sum_r t_r)$, a multiple of $n$. Applying the projection to a minimum-degree certificate proves the last statement. Averaging further over permutations of the $R$ labels is also valid in characteristic zero and gives the orbit representation used by the computation.
\end{proof}

\subsection{An explicit boundary certificate}

Consider the tensor $S\in\C^2\otimes\C^2\otimes\C^2$ with slices $I_2,J_2$. With indices $\alpha,i,j\in\{0,1\}$, let
\[
 F_{\alpha i j}=\sum_{r=1}^{2}a_{r,\alpha}b_{r,i}c_{r,j}-S_{\alpha i j}.
\]
For a monomial $p$ in the six symbols $a_0,a_1,b_0,b_1,c_0,c_1$, write $p^{(r)}$ for its evaluation at label $r$, and define
\[
                  A(p,q)=\frac{p^{(1)}q^{(2)}+p^{(2)}q^{(1)}}{2}.
\]
This convention also applies when $p=q$.

\begin{lemma}[Degree-9 boundary identity]\label{lem:base}
The sum of $\kappa A(p,q)F_{\alpha i j}$ over the rows of \cref{tab:base-certificate} equals one. Every multiplier is label-balanced and has degree at most six. Thus $\delta_2(S)=9$.
\end{lemma}

\begin{table}[htbp]
\centering\small
\begin{tabular}{@{}rlll@{}}
\toprule
$\kappa$ & $p$ & $q$ & $(\alpha,i,j)$ \\
\midrule
$-1$ & $1$ & $1$ & $(1,0,1)$ \\
$4$ & $1$ & $a_1b_1c_1$ & $(0,0,1)$ \\
$-2$ & $1$ & $a_1b_0c_1$ & $(0,1,1)$ \\
$2$ & $1$ & $a_0a_1b_0b_1c_1^2$ & $(1,0,1)$ \\
$-2$ & $1$ & $a_0a_1b_0^2c_1^2$ & $(0,1,0)$ \\
$2$ & $1$ & $a_0a_1b_0^2c_1^2$ & $(1,0,0)$ \\
$-2$ & $1$ & $a_0a_1b_0^2c_1^2$ & $(1,1,1)$ \\
$2$ & $1$ & $a_0a_1b_0^2c_0c_1$ & $(0,1,1)$ \\
$-2$ & $1$ & $a_0a_1b_0^2c_0c_1$ & $(1,0,1)$ \\
$4$ & $1$ & $a_0^2b_0^2c_1^2$ & $(1,1,0)$ \\
$-4$ & $1$ & $a_0^2b_0^2c_0c_1$ & $(1,1,1)$ \\
$2$ & $a_1b_1c_1$ & $a_1b_0c_1$ & $(0,0,1)$ \\
$4$ & $a_1b_1c_1$ & $a_0b_0c_1$ & $(0,0,0)$ \\
$-4$ & $a_1b_1c_0$ & $a_0b_0c_1$ & $(0,0,1)$ \\
$2$ & $a_1b_0c_1$ & $a_1b_0c_1$ & $(0,0,0)$ \\
$-2$ & $a_1b_0c_1$ & $a_1b_0c_1$ & $(0,1,1)$ \\
$-2$ & $a_1b_0c_1$ & $a_1b_0c_0$ & $(0,0,1)$ \\
$-2$ & $a_1b_0c_1$ & $a_0b_1c_1$ & $(0,0,0)$ \\
$2$ & $a_1b_0c_1$ & $a_0b_1c_0$ & $(0,0,1)$ \\
\bottomrule
\end{tabular}
\caption{The complete boundary certificate. Each row contributes $\kappa A(p,q)F_{\alpha i j}$; their sum is the constant one.}
\label{tab:base-certificate}
\end{table}

\begin{proof}
Substitute $S_{0ij}=\mathbf1_{i=j}$ and $S_{1ij}=\mathbf1_{(i,j)=(0,1)}$ in the displayed sum and expand. All nonconstant coefficients cancel and the constant coefficient is one. The table specifies the entire rational identity, so this verification requires only polynomial multiplication and collection; the companion also implements that expansion directly. The mode degrees in each displayed $p$ and $q$ agree, and each $A(p,q)$ has degree at most six. The product degree is at most nine. A certificate proves inconsistency, and \cref{lem:lower} supplies the matching lower bound.
\end{proof}

Swapping $e_0$ and $e_1$ in the third mode sends $S$ to $W_3$. This is a constant linear change of variables, so the same certificate, with relabeled coordinates and equations, applies to $W_3$ with the same degrees and balance.

\subsection{A degree-controlled transfer lemma}

\begin{lemma}[Balanced transfer]\label{lem:transfer}
Let a $k$-mode tensor $S$ have a CP-$R$ unit certificate whose multipliers are label-balanced and have degree at most $k(q-1)$. Let $n\ge k$, and suppose an $n$-mode tensor $T$ maps to $S$ by constant linear maps on $k$ retained modes and constant linear functionals on the remaining modes. Then the original CP-$R$ system for $T$ has a unit certificate of product degree at most $nq$.
\end{lemma}

\begin{proof}
Permute modes so that the retained modes are $1,\ldots,k$. Write $A_i:V_i\to U_i$ for their linear maps and $\ell_i:V_i\to\C$ for the other functionals. For each summand label, substitute into the base parameters
\[
 y_{r,1}=A_1x_{r,1}\prod_{i=k+1}^{n}\ell_i(x_{r,i}),
 \qquad y_{r,j}=A_jx_{r,j}\quad(2\le j\le k).
\]
The first vector has degree at most $n-k+1$ and each other vector has degree at most one. A balanced multiplier monomial with degree $t_r$ in each of the $k$ modes at label $r$ becomes a polynomial of degree at most
\[
 \sum_r\bigl((n-k+1)t_r+(k-1)t_r\bigr)=n\sum_r t_r.
\]
Hence every substituted multiplier has degree at most $n(q-1)$.

Let $L=(\bigotimes_{i=1}^k A_i)\otimes(\bigotimes_{i=k+1}^n\ell_i)$ be the resulting constant linear map on tensor entries. By construction, $Q^{S}(y)=L(Q^{T}(x))$ and $S=L(T)$. Therefore each substituted base equation is a constant linear combination of the original equations:
\[
 F^S_K(y(x))=\sum_J L_{KJ}F^T_J(x).
\]
Substitution in the base unit identity gives
\[
 1=\sum_K h_K(y(x))\sum_J L_{KJ}F^T_J(x).
\]
Each multiplier has degree at most $n(q-1)$ and each original equation degree $n$, proving the bound. The construction is a polynomial identity even when some contracted parameter forms vanish.
\end{proof}

\begin{corollary}[Normalized certificate degree under contraction]\label{cor:degree-transfer}
Let $R\ge1$ and $n\ge k\ge1$. Let $T$ be an $n$-mode tensor and $S$ a $k$-mode tensor over $\C$, with $S$ obtained from $T$ by constant linear maps on $k$ retained modes and constant linear functionals on the other modes. If $\delta_R(S)<\infty$, then
\begin{equation}\label{eq:degree-transfer}
 \delta_R(T)\le\frac{n}{k}\,\delta_R(S),
 \qquad\text{equivalently}\qquad
 \frac{\delta_R(T)}{n}\le\frac{\delta_R(S)}{k}.
\end{equation}
Both degrees are measured in their respective original CP-$R$ parameter systems.
\end{corollary}

\begin{proof}
Put $D=\delta_R(S)$ and choose a certificate attaining this finite minimum. Applied to the $k$-mode system, \cref{lem:balance} gives a balanced certificate of degree at most $D$ and shows that $D=kq$ for an integer $q$. Its multipliers have degree at most $D-k=k(q-1)$, because every original equation has degree $k$. \Cref{lem:transfer} therefore supplies a certificate for $T$ of degree at most $nq=(n/k)D$. In particular the transferred system is inconsistent as well. No separate hypothesis that the chosen certificate is already balanced is needed.
\end{proof}

\begin{corollary}\label{cor:wn}
For every $n\ge3$, the tensor $W_n$ in \cref{eq:wn} satisfies $\delta_2(W_n)=3n$.
\end{corollary}

\begin{proof}
Contract modes $4,\ldots,n$ by $e_0^*$. The result is $W_3$. By \cref{lem:base}, $\delta_2(W_3)=9$. Applying \cref{cor:degree-transfer} with $k=3$ and $R=2$ gives an upper certificate of degree $(n/3)\cdot9=3n$. The certificate proves the fiber is empty, and \cref{lem:lower} gives the opposite inequality. The flattening formula and degeneration in \cref{sec:boundary} show that every flattening has rank at most two and the border rank is two. No secant classification is needed for this corollary.
\end{proof}

\subsection{Proof of the full two-term theorem}

\begin{proof}[Proof of \cref{thm:main}]
The lower bound $3n$ is \cref{lem:lower} with $R=2$. If a flattening has rank greater than two, \cref{lem:minor} gives the matching upper bound. This covers $n=2$: a matrix of rank greater than two has a nonzero $3\times3$ minor, so the two lemmas give $\delta_2(T)=6=3n$. We may therefore assume $n\ge3$. It remains to consider tensors for which every bipartition flattening has rank at most two.

The set-theoretic flattening theorem of Landsberg and Manivel \cite[arXiv version, Theorem~5.1]{landsbergmanivel} places such a tensor in the second secant variety of the Segre variety. Raicu proves the stronger ideal-theoretic statement \cite[Theorem~4.1 and Corollary~4.2]{raicu}. Write $X$ for this Segre variety. Points of $\sigma_2(X)\setminus\tau(X)$ have rank two \cite[arXiv version, Corollary~1]{ballicobernardi}. Hence $\rank(T)>2$ forces $[T]\in\tau(X)$. By \cite[arXiv version, Notation~1]{ballicobernardi}, $\tau(X)$ is the union of tangent spaces, so $T$ has the form \cref{eq:tangent-main} with nonzero $v_i$ and appropriate $u_i$.

Absorb components of $u_i$ parallel to $v_i$ into $\lambda$. Inactive modes now have $u_i=0$. There must be at least three active modes: with at most two, factoring out passive vectors leaves a vector or a matrix supported on a two-dimensional space in each active mode, of rank at most two. Select three active modes. On each, choose a linear map sending $v_i$ to $e_0$ and $u_i$ to $e_1$. On every unselected mode, choose a functional taking $v_i$ to one and $u_i$ to zero. The resulting three-mode tensor is
\[
                 \lambda e_0\otimes e_0\otimes e_0+W_3.
\]
In the first retained mode, the linear shear fixing $e_0$ and sending $e_1$ to $e_1-\lambda e_0$ removes the pure term. The target therefore maps to $W_3$ by maps of the form required in \cref{lem:transfer}. With the degree-9 certificate of \cref{lem:base}, this gives the upper bound $3n$.

Dimension-one factors may be suppressed when applying the classical geometric statement, since they do not change tensor rank or flattening rank. They remain among the original parameter modes and are contracted in the transfer, so each is still charged in the factor $n$. If fewer than three nonscalar active modes remain in this case, the rank would be at most two, already excluded. This also accounts for the endpoint dimensions.
\end{proof}

\subsection{Uniform certificates for nilpotent pencils}\label{app:pencil}

We prove \cref{thm:pencil} constructively. For a polynomial $p$, write $p\in\mathcal I_{\le d}$ if it has an expression as a linear combination of the original coefficient equations in which every multiplier--equation product has degree at most $d$. Sums preserve the larger bound, and multiplication by a degree-$e$ polynomial raises it by at most $e$.

First choose a constant matrix $H$ satisfying $NH-HN=N$. In Jordan form, take diagonal entries $1,2,\ldots$ successively within each block and then conjugate $H$ back. Introduce square matrices $B,C$ with columns $b_r,c_r$, and diagonal matrices
\[
 A=\diag(a_1,\ldots,a_m),\qquad D=\diag(d_1,\ldots,d_m).
\]
Set
\[
 X=BAC^{\mathsf T},\quad Y=BDC^{\mathsf T},\quad
 F=X-I_m,\quad G=Y-N,
\]
and
\[
 \beta=\det B,\quad \gamma=\det C,\quad U=\beta\gamma,
 \quad P=\prod_{r=1}^m a_r.
\]
The entries of $F,G$ are exactly the original cubic generators. Adjugate identities give
\begin{align*}
 L&=(NF-G)\adj(C^{\mathsf T})=\gamma(NBA-BD),\\
 R'&=\adj(B)(FN-G)=\beta(AC^{\mathsf T}N-DC^{\mathsf T}).
\end{align*}
Their entries belong to $\mathcal I_{\le m+2}$. Write $L_r$ for column $r$ of $L$ and $R'_r$ for row $r$ of $R'$. Then
\begin{align*}
 S_r&=\gamma R'_rHb_r-\beta c_r^{\mathsf T}HL_r
      =Ua_r c_r^{\mathsf T}(NH-HN)b_r
      =Ua_r c_r^{\mathsf T}Nb_r,\\
 T_r&=\beta c_r^{\mathsf T}L_r
      =U\bigl(a_r c_r^{\mathsf T}Nb_r-d_r c_r^{\mathsf T}b_r\bigr).
\end{align*}
Both belong to $\mathcal I_{\le2m+3}$. Setting $z_r=c_r^{\mathsf T}b_r$ gives
\[
 W_r=S_r-T_r=Ud_rz_r\in\mathcal I_{\le2m+3}.
\]
Also
\[
 V_r=(\adj(B)FB)_{rr}=\beta(a_rz_r-1)
       \in\mathcal I_{\le m+3}.
\]
Consequently,
\[
 K_r=a_rW_r-d_r\gamma V_r=Ud_r\in\mathcal I_{\le2m+4}.
\]
Choose fixed indices $i,j$ with $N_{ij}\ne0$. The identity
\[
 Z=\frac{\sum_r(b_r)_i(c_r)_jK_r-UG_{ij}}{N_{ij}}=U
\]
places $U$ in $\mathcal I_{\le2m+6}$. The denominator is a nonzero field constant. Multiplying by $P$ yields $PZ=\det X\in\mathcal I_{\le3m+6}$. Finally, determinant telescoping gives $\det X-1\in\mathcal I_{\le3m}$. Subtraction produces the unit certificate
\[
                      1=PZ-(\det X-1)\in\mathcal I_{\le3m+6}.
\]
Every displayed intermediate expression has an explicit combination of the original equations, so substitution gives all final multipliers, including at singular parameter matrices.

\begin{table}[htbp]
\centering\small
\begin{tabular}{@{}ll@{}}
\toprule
Derived expression & Maximum original product degree \\
\midrule
Entries of $L,R'$ & $m+2$ \\
$S_r,T_r,W_r$ & $2m+3$ \\
$V_r$ & $m+3$ \\
$K_r=Ud_r$ & $2m+4$ \\
$Z=U$ & $2m+6$ \\
$PZ=\det X$ & $3m+6$ \\
$\det X-1$ & $3m$ \\
Unit identity & $3m+6$ \\
\bottomrule
\end{tabular}
\caption{Degree ledger for the nilpotent-pencil certificate. Derived equations are charged through their representations in the original cubic generators.}
\label{tab:degree-ledger}
\end{table}

For completeness, $(I_m,N)$ has no rank-$m$ decomposition. If $I_m=BAC^{\mathsf T}$, all three square factors are invertible. Then $N=BDC^{\mathsf T}=BDA^{-1}B^{-1}$ is diagonalizable, impossible for a nonzero nilpotent matrix. In Jordan coordinates, the perturbations $N+\varepsilon\diag(1,\ldots,m)$ have distinct eigenvalues for $\varepsilon\ne0$; the corresponding pencils have rank $m$ and converge to $(I_m,N)$. The identity slice supplies the lower border-rank bound $m$, and the three flattening ranks are $2,m,m$.

Apply \cref{lem:lower} with order three and $R=m$ to get the lower degree $3m+3$. \Cref{lem:balance} makes the minimum degree divisible by three. Together with the constructive upper bound, this proves \cref{thm:pencil}. The rational certificates at $m=2,3,4$ select the lower endpoint for single Jordan blocks; the uniform theorem retains both possibilities.

\section{Records, artifacts, and replay}\label{app:records}

\subsection{An interoperable record specification}

The following fields specify the information an RRE implementation should expose. The case-study files implement these concepts through a charter, typed graph records, pinned evidence, event records, and version deltas. The underlying atlas's extraction schema is described in \cref{app:complexity-graphs}.

\begin{longtable}{@{}p{2.0cm}p{4.0cm}p{7.8cm}@{}}
\toprule
Record & Core fields & Semantic requirement \\
\midrule
\endhead
Claim & ID, revision, statement, role, scope, source & Quantifiers and parameter regime belong to the statement; reconstructed text and source evidence remain distinguishable. \\
Inference & ID, premise IDs, conclusion, scope, justification & All premises are jointly required. Alternative proofs and conditional branches receive separate records. \\
Target & ID, parent, statement, objective link, bounds, status & Success must meet the stated quantitative requirement; partial results remain separately usable. \\
Objective & ID, origin, rationale, acceptance conditions, revision & Preserve the original objective and any prospectively recorded family changes. \\
Evidence & Type, location, hash, inputs, environment & Distinguish exact computations, source assertions, written proofs, and formal certificates. \\
Review & Reviewer role, coverage, method, findings & Report the actual scope of checking. \\
Event & ID, time, actor, target, action, inputs, outcome, change & Separate the actor of an action from the authorship of its imported sources. \\
Assessment & Result, significance, rationale, literature comparison & Assess resolved obligations separately from proof status. \\
Run & Mode, objective, initial graph, sources, resources, stop reason & Retain acceptance conditions and the actual stopping decision. \\
Strategy revision & Method versions, evaluation tasks, costs, decision & Evaluate changes to the research procedure separately from its mathematical outcomes. \\
\bottomrule
\end{longtable}

A node ID remains stable while a version identifies its exact content. A source anchor records a file or public source version, its location, and a hash where available. Graph evidence in this run pins local mathematical notes, certificates, checks, and reading records. Public references specify the consulted versions; a hash of a local reading note is not presented as a hash of the cited PDF.

For example, the exact two-term theorem has an exhaustive-case inference. Its inputs are the lower bound, the geometric dichotomy, and two conditional upper-bound routes. The boundary route depends on the projection to $W_3$, the base certificate, and balanced transfer. The separate $W_n$ inference uses the explicit family contraction and does not import the general geometric classification. Both traces retain the original-parameter degree convention. The historical event that proposed transfer is linked to these claims as motivation rather than inserted as a mathematical premise. The normalized inequality in \cref{cor:degree-transfer} and the matrix endpoint of \cref{thm:main} make further consequences of the recorded lemmas explicit. The six graph versions retain their original statements and scopes as the exploration history.

\subsection{Artifacts and portable replay}

The companion contains the six component graphs and their deltas, the charter, the source pins, 44 event records, the mathematical notes, exact certificates, and computational reports. A source manifest records the included bytes and their SHA-256 hashes. The portable replay program checks the saved rational certificates directly in the original variables and replays the transferred $W_n$ identities. It also validates graph identifiers, scopes, acyclic proof dependencies, version links, and event hashes. The original research utilities and saved reports are retained for comparison; replay writes its own report.

\begin{table}[htbp]
\centering\small
\begin{tabularx}{\linewidth}{@{}p{5.0cm}Y@{}}
\toprule
Artifact & What it makes reproducible or inspectable \\
\midrule
Charter and initial source pins & Prospective objective, degree convention, inherited inputs \\
Six graphs and version deltas & Incremental additions, revisions, and retained open questions \\
Event history and replay index & Targets, actions, outcomes, actor attribution, and artifact versions \\
Mathematical notes and certificate table & Uniform arguments and the complete degree-9 identity \\
Rational certificates for three single-block pencils & Raw-variable unit identities of degrees 9, 12, and 15 \\
Transfer and contraction checks & All-order construction tested on explicit finite inputs \\
Uniform commutator checks & Symbolic identities and singular-parameter controls \\
Prior-work comparison & Classical inputs and the specific certificate-degree contribution \\
Viewer, captures, and figure map & Connection between displayed claims and saved graph versions \\
Dedicated discovery exports & Prerequisite paths and evidence for each principal theorem \\
\bottomrule
\end{tabularx}
\caption{Principal artifacts of the certificate-degree case study. Exact filenames and replay commands are supplied in the companion documentation.}
\end{table}

The rational certificate files are replayed by expanding each normalized orbit monomial into original parameter monomials, multiplying by its original coefficient equation, and collecting coefficients over $\Q$. The result must be exactly the constant one, and the largest product degree is checked. This verification uses the certificate rather than repeating the elimination search. The appendix's 19-row table is also compared against the certificate file, so the printed mathematical object and its machine-readable counterpart agree.

The all-order transfer is proved in \cref{lem:transfer}; its finite replay uses the same base coefficients with the contracted parameter factors retained. Source-hash validation checks which recorded artifact version supports each claim. Mathematical checking and provenance checking produce separate outcomes. Replaying the saved artifacts does not require restarting the original research agent or regenerating its target choices.

\subsection{Event-level effects}\label{app:event-effects}

The table groups all 44 saved records by their effect. The raw display IDs are preserved. Two concurrent computation-start records both used E0022; their distinct event hashes identify them uniquely, and the replay index assigns unique sequential identities. No mathematical result depends on the duplicated display number.

\small
\begin{longtable}{@{}p{2.7cm}p{11.3cm}@{}}
\toprule
Recorded events & Effect on the investigation and graph \\
\midrule
\endhead
E0001--E0004 & Adopt the objective and initial targets, seal v001, and publish its atlas component. \\
E0005--E0008 & Perform the first degree tests, verify the matrix-pencil baseline, extend the lower functional to unequal dimensions, and prove complete symmetry reduction. \\
E0009--E0012 & Generate the all-size sharp-bound and target-grading questions, publish v002, and replay the rational certificates by raw expansion. \\
E0013--E0015 & Use target grading, begin the rational lift of the degree-15 test, and save its exact certificate and expansion. \\
E0016--E0019 & Generate the uniform commutator target, check its identities, record the all-$m$ theorem, and generate the remaining endpoint question. \\
E0020--E0021 & Seal and publish v003 with the uniform theorem and its proof obligations. \\
E0022 (two records), E0024--E0026 & Start two multiple-block screens, save their distinct outcomes, and preserve and reconcile the concurrent event-log fork. \\
E0027--E0031 & Publish v004, correct an export path-length issue, export the pencil theorem, and save the open exact-endpoint checkpoint. \\
E0032--E0033 & Clarify the acceptance policy without weakening the mathematical objective; prospectively generate the balanced-transfer and full CP-2 target. \\
E0034--E0036 & Replay transfer and contraction maps, then save the all-order exact theorem and its $W_n$ specialization. \\
E0037--E0038 & Seal and publish v005, including the classical imported supports and the two conditional routes. \\
E0039--E0042 & Audit the theorem's scope, degree accounting, and case coverage; assess the original criteria; seal and publish v006. \\
E0043--E0044 & Export the exact theorem's prerequisite graph and preserve the completed checkpoint and replay outline. \\
\bottomrule
\end{longtable}
\normalsize

The event fork preserves both original records and their common predecessor. A reconciliation record explains the fork, and subsequent writes use serialization. Playback follows the recorded file order with unique event hashes; the raw log remains unchanged. This is an explicit part of the provenance rather than a reconstructed single chain with an event silently removed.

The six graph publications are actual saved revisions. Each summarizes the claims admitted by that checkpoint; it is not an invented graph state for every intervening action. The final proof can therefore be read in dependency order, while the events preserve the distinct order in which targets, checks, and proofs were recorded.

\subsection{Visualization and evidence links}

The companion supplies the original graph data embedded in offline viewer pages, one per version. Selecting an inference exposes all its premises and its justification; selecting a claim exposes its statement, attached scopes, status, and evidence. Prerequisite traversal follows producing inferences and attached scopes. Navigation links are optional and do not participate in that traversal.

The printed overview uses all six saved graphs, while the inspector image shows a selected record and its evidence. The browser validation report records graph hashes and counts and checks the prerequisite closures of the two-term theorem and $W_n$. The capture routine also checks that displayed node text stays within its cards. The timeline is an explanatory diagram keyed to the same versions and events; its arrows describe the emergence of research targets rather than proof implications.

The versioned viewer extends the earlier graph-comparison paradigm by providing a separate view for each of the six saved revisions. The companion pairs these views with the corresponding graph data and evidence.

\clearpage

\section{Construction of the complexity reasoning graph collection}\label{app:complexity-graphs}

This appendix describes the complexity atlas through October 31, 2026, including the later exploration component from which the certificate-degree investigation drew its inputs. The October 25 extraction is the base of this evolving collection. It separates three objects: the extracted \emph{complexity atlas}, the selected \emph{working graph} used for the certificate-degree investigation, and the \emph{increment} produced by that investigation. The atlas implements source reconstruction; the run illustrates research that uses and extends selected reasoning. Neither the collection size nor its extraction status measures the amount of independently verified mathematics.

\subsection{Sources, snapshots, and units of accounting}

The atlas is a local, dated collection of original complexity research notes, manuscript sections, arguments, research histories, static code contracts, and stored computational results. Its October 25, 2026 completion record covers five local research collections spanning complexity notes, problem directions, prior exploratory records, and focused investigations. Source location records provenance, not authorship. Separate research streams retain their original files and hypotheses.

\Cref{tab:atlas-evolution} distinguishes the original extraction from the atlas available when the present case study began and the atlas after its final integration. Between the extraction snapshot and the new run, subsequent components added 4,578 node records and 2,667 AND inferences. The case study then contributed a component of 68 nodes and 34 AND inferences.

\begin{table}[htbp]
\centering\small
\begin{tabularx}{\linewidth}{@{}Yrr@{}}
\toprule
Dated atlas state & Nodes & AND inferences \\
\midrule
October 25: original extraction & 58,916 & 19,516 \\
October 31: before the case study & 63,494 & 22,183 \\
October 31: after the case study & 63,562 & 22,217 \\
\bottomrule
\end{tabularx}
\caption{Cumulative atlas counts through October 31, 2026. Each row is a complete dated state, not an additional component. Only the registered version of each component is counted, and references to existing base nodes are not counted again.}
\label{tab:atlas-evolution}
\end{table}

The document and batch counts below describe the original extraction; later research components are accounted for by their graph records. The extraction denominator excludes 49 adjacent collections, downloaded libraries, Git internals, and generated infrastructure. External theorem statements may appear as imports; their full proofs are not thereby extracted. Completion means that the declared snapshot's in-scope files received the recorded extraction or documentary treatment, with no pending or stale in-scope files in that audit. It does not describe later additions or every research subject in the workspace.

\begin{table}[htbp]
\centering\small
\begin{tabularx}{\linewidth}{@{}Yrp{5.8cm}@{}}
\toprule
Unit & Count & Interpretation \\
\midrule
Primary documents & 1,874 & Source files or sections, not necessarily papers \\
Reconciled history documents & 855 & Earlier records related to later argument records \\
Extraction batches & 4,200 & Units of annotation and accounting \\
Mathematical argument groups & 7,738 & Local groups of claims, scopes, and steps \\
History and support groups & 17,406 & Documentary, review, and supporting records \\
Nodes & 58,916 & Multiple semantic roles, not only theorems \\
AND inferences & 19,516 & Authored joint-premise derivations \\
Cross-batch navigation links & 15,515 & Connections that do not assert entailment \\
Distinct source-file pins & 12,743 & Primary and supporting file references \\
\bottomrule
\end{tabularx}
\caption{The complexity atlas snapshot completed on October 25, 2026. Counts are different units and should not be added as if they measured independent results.}
\label{tab:atlas-counts}
\end{table}

The 4,200 batches include 1,691 with local inferences and 2,509 without them. Definitions, imported contracts, historical records, and file-accounting groups can legitimately have no local derivation. Of the nodes, 19,308 have role \code{derived_claim}, 18,207 \code{review_record}, 967 \code{open_obligation}, and 244 \code{counterexample}. These roles describe recorded content. In particular, a source-reported derived claim is not automatically a theorem accepted by RRE.

\subsection{Semantic extraction and assembly}

The annotations were drafted by the AI assistant through reading and explicit semantic authoring. The compiler assembles those annotations and checks their structure; it does not autonomously reconstruct arguments from unannotated prose. The procedure is:

\begin{enumerate}[leftmargin=1.7em]
\item \textbf{Inventory and pin the sources.} Record file paths, SHA-256 hashes, source roles, and immutable snapshots. Separate mathematical documents from histories, supporting programs, saved results, and excluded infrastructure.
\item \textbf{Identify argument units.} Select a conclusion and reconstruct its supporting steps from the source. Paragraph order, common vocabulary, and citation links alone do not establish a dependency.
\item \textbf{Expose the hypotheses.} Record the field, domain, quantifier order, fixed and growing parameters, degree and support bounds, exceptional cases, and distinctions such as exact versus border statements. A proposed bridge remains conditional when its hypotheses have not been shown for the intended circuit or target.
\item \textbf{Author claims and AND steps.} Give each claim a role, source passage, and scope. List jointly required premises explicitly. Keep definitions, external contracts, computational assertions, counterexamples, and unresolved obligations distinct.
\item \textbf{Reconcile and connect.} Relate historical attempts to later arguments without upgrading old confidence statements into proofs. Add cross-batch navigation for relevant tools, obstructions, and possible transfers; retain the proof obligations for using those connections.
\item \textbf{Compile and check.} Assemble batch graphs, propagate recorded scope, check identifiers and evidence anchors, test prerequisite traces, account for source coverage, and export the navigable collection.
\end{enumerate}

Source programs were read statically during this extraction. A saved exact computation retains the status of a computation reported by its source unless a separately identified replay was performed. The fresh checks of the certificate-degree investigation in \cref{sec:case} are a different operation from extracting the atlas.

\subsection{Implemented records and scope propagation}

The implemented atlas schema is narrower than the proposed RRE record specification in \cref{app:records}. Its principal fields are summarized below; the companion sample includes the actual JSON schema and representative records.

\begin{table}[htbp]
\centering\small
\begin{tabularx}{\linewidth}{@{}p{2.2cm}Y@{}}
\toprule
Record & Implemented information \\
\midrule
Source & Path, full-file SHA-256, snapshot path, source role \\
Evidence anchor & Source ID, inclusive line interval, exact decoded passage, passage SHA-256 \\
Node & Stable ID, authored title, verbatim source passages, role, source-reported status, scope IDs, review fields, planner eligibility \\
Inference & ID, premise IDs, conclusion, scope IDs, effective premises, reasoning mode, evidence; explicit assumption discharge where needed \\
Argument group & Nodes, targets, inferences, scope, and interpretation of conditional hypotheses \\
Navigation relation & Endpoints, relationship kind, explanatory note, and \code{asserts_entailment=false} \\
\bottomrule
\end{tabularx}
\caption{Existing atlas records. An authored summary and its verbatim evidence remain separate.}
\end{table}

For an inference, the compiler begins with its explicit premises, its own scope, and the conclusion's attached scope. It then repeatedly adds scopes attached to these nodes to produce \code{effective_premises}. This prevents a recorded hypothesis from disappearing merely because a viewer displays only the immediate premise list. It cannot recover a hypothesis that the annotation omitted. When an argument discharges a local assumption, that event is recorded separately; the assumption's appearance in the historical prerequisite trace does not make it an undischarged hypothesis of the conclusion.

All nodes and inference records in the original extraction have \code{planner_eligible=false}. Their review fields distinguish source grounding from mathematical certification and say that human review was not assessed in this import. RRE must therefore select, interpret, and assess a proposed input before treating it as supported, or retain it explicitly as a conditional source dependency. Importing a graph does not silently admit all its assertions into a proof engine.

\subsection{A concrete conditional argument}

Batch CB4173 illustrates why this representation matters. Its source is a dated transfer-and-proofs note in the atlas integration materials, lines 129--147. The scope fixes source coefficient polynomials $Q_i$ with identically zero off-support coefficients, target polynomials $\mu_i$, and the \emph{full} space of coefficient monomials of degree at most $D$. Let
\[
 A_D(P)=P(Q),\qquad T_D(P)=P(\mu).
\]
The node \code{CB4173:product-blind} is the hypothesis $\ker A_D\subseteq\ker T_D$. Under that hypothesis,
\[
 \Phi_D(P(Q))=P(\mu),\qquad \deg P\le D,
\]
is well defined and linear: two representatives with the same source substitution differ by an element of $\ker A_D$, which the hypothesis places in $\ker T_D$. If $\deg P\le a$, $\deg R\le b$, and $a+b\le D$, substitution gives
\[
 \Phi_D\bigl(P(Q)R(Q)\bigr)
   =\Phi_D(P(Q))\Phi_D(R(Q)).
\]
The product is available because the full degree-bounded space contains $PR$.

The recorded step \code{CB4173:multiplicative-infer} has the three premises \code{product-scope}, \code{product-blind}, and \code{truncated-map}, all in the CB4173 namespace. Thus a trace exposes the full-dictionary scope and the kernel-inclusion hypothesis. It does not present multiplication compatibility as an unconditional consequence of a sparse linear reconstruction. Separate nodes retain the cost of the full dictionary and the unresolved obligation to obtain an affordable extension for actual circuit sources. This is an example of an existing conditional argument, not an additional discovery of the present investigation. The short derivation and its source anchors were checked during preparation of this appendix; no broader novelty or human-review claim follows.

\subsection{From the atlas to an evolving working graph}

The case study starts from the October 31 atlas, not solely from the October 25 extraction. Its immediate source is revision 241 of the preceding exploration component, RRE003, with 4,108 nodes and 2,362 AND inferences. The new run pins this graph and its supporting note, then selects two results: the distinct-label lower functional and the matching certificate when a flattening has rank greater than the proposed tensor rank. Their canonical records are \code{RRE003:cp119-lower} and \code{RRE003:cp119-exact}. These become explicit inherited inputs in the new component, RRE005; their hypotheses are checked for the new family. Thus the case study reuses results of an earlier exploration already incorporated into the atlas. Earlier investigations of coefficient reconstruction and parameter normalization, related to scaling reparametrizations such as \cite{meshkat}, also remain available in the collection.

The new component starts with eight records identifying its objective, parameter convention, inherited results, and boundary target. Successive publications add proofs, finite certificates, generated targets, and classical imports. Its six revisions contain respectively 8, 19, 43, 46, 66, and 68 nodes, and 0, 7, 22, 22, 33, and 34 AND inferences. The final component includes its eight initial records, so growth after initialization is 60 nodes and 34 inferences. Each publication stores the complete component, a delta from its predecessor, and validation results, then advances only this component's atlas pointer. Other components retain their own versions.

At the final integration on October 31, the registry reaches the last row of \cref{tab:atlas-evolution}, with 17,663 navigation links across the base and subsequent components. This navigation total includes both local and cross-component relations; the 15,515 links in \cref{tab:atlas-counts} count only the original cross-batch relations. The 68-node component is the object analyzed in the case study, while the larger atlas supplies the research context and selected inherited inputs. Two dedicated discovery exports retain the full prerequisite paths for the nilpotent-pencil theorem and the exact two-term theorem, with copies of their hash-pinned evidence. This permits a separate visualization process to inspect a result without copying the entire atlas.

The collection viewer serves corpus navigation; the certificate-study companion serves the selected investigation. The latter offers a saved view for each version, complete recorded prerequisite traces, AND-step inspection, and source evidence. The printed overview in \cref{fig:certificate-versions} relates the beginning and end of this component. Its timeline in \cref{fig:discovery-timeline} describes research actions, while the theorem view in \cref{fig:certificate-proof} describes implication.

\subsection{Validation, limitations, and companion material}

The dated atlas audit reports 105,294 evidence anchors checked, 2,725 prerequisite regressions, and automated browser checks of 25,144 argument groups and their expanded traces. These checks concern source alignment, expected graph behavior, and interface operation. Nonblank-line coverage also includes documentary classifications and broad source spans; it is not a measure of semantic accuracy. Neither automated coverage nor a clean graph certifies every inference or establishes novelty.

For the original extraction appendix, the full graph's SHA-256 was checked against the completion and statistics reports. Batch totals, node roles and statuses, anchor counts, and planner-ineligibility flags were recounted from all 4,200 batch graphs. The sources, snapshots, and evidence anchors of CB291 and CB4173 were checked again. These read-only checks did not rerun all historical regressions, browser tests, or source computations. The new component has separate source-hash, dependency, exact-certificate, and viewer checks, documented in \cref{app:records}.

For the October 31 accounting, the base graph and all five registered components were checked against their recorded SHA-256 hashes. Node and inference counts were recomputed, repeated base references were excluded, and the totals were matched to the registry saved before the case study and the registry after its final integration. The two direct imports were traced to the pinned prior graph and proof note. The companion accounting report retains these source pins, the two registry snapshots, and the recount program.

The local atlas is organized as a compiled semantic graph, with batch graphs and authored specifications stored separately. The paper package supplies a representative atlas sample containing the extraction documentation, schema, dated reports, and two batches with their sources. The accompanying source manifest records paths and SHA-256 hashes, and the audit report separates newly performed checks from historical reported checks. The full atlas is not bundled with this draft. This sample documents the method and its limits without implying that the entire collection has been publicly released.

\end{document}